\RequirePackage{fix-cm}
\documentclass[twocolumn]{svjour3}    % twocolumn
\pdfoutput=1

\usepackage{graphicx}
\usepackage{natbib}

\usepackage{url}            % simple URL typesetting
\usepackage{booktabs}       % professional-quality tables
\usepackage{amsfonts}       % blackboard math symbols
\usepackage{nicefrac}       % compact symbols for 1/2, etc.
\usepackage{microtype}      % microtypography
\usepackage{graphicx}
\usepackage{enumitem}
\usepackage{bm}
\usepackage{nccmath}
\usepackage{boldline}
\usepackage{booktabs}        % To thicken table lines

\usepackage{algorithm}
\usepackage[noend]{algcompatible}

\usepackage{capt-of}
\usepackage{color,soul}
\usepackage[pagebackref=false,breaklinks=true,colorlinks,bookmarks=false,urlcolor=blue, citecolor=blue]{hyperref}

\usepackage{times}
\usepackage{epsfig}
\usepackage{multirow}
\usepackage{xcolor}
\usepackage{amsmath}
\usepackage{amssymb}
\usepackage{wrapfig}
\usepackage{comment}
\usepackage{textcomp}
\usepackage{bm}
\usepackage{soul}

\usepackage{stfloats}
\usepackage{cuted}
\usepackage{capt-of}
\usepackage{marvosym}
\usepackage{graphicx}
\usepackage{booktabs}
\usepackage{enumitem}
\usepackage{amsfonts}
\usepackage{bbm}

\usepackage{makecell}
\usepackage{subfigure}
\usepackage{arydshln}
\usepackage{ulem}

\def\tb#1{\textbf{#1}}
\def\cm#1{\checkmark}

\newcommand{\etal}{\textit{et al}. }

\useunder{\underline}{\ul}{}

\usepackage{tikz}

\begin{document}

\title{A Closer Look at  Few-Shot 3D Point Cloud Classification}

\author{Chuangguan~Ye$^{\ast}$ \and Hongyuan~Zhu$^{\ast}$ \and Bo~Zhang \and Tao~Chen$^{\dagger}$}

\institute{Chuangguan~Ye  \at
         School of Information Science and Technology,\\ 
         Fudan University, \\ 
         \email{cgye19@fudan.edu.cn}
         \and
         Hongyuan~Zhu \at
         Institute for Infocomm Research, \\
         A*STAR,\\
         \email{hongyuanzhu.cn@gmail.com}
         \and
         Bo~Zhang \at
         Shanghai AI Laboratory, \\
         \email{zhangbo@pjlab.org.cn}
         \and
         Tao~Chen \at
         School of Information Science and Technology,\\ 
         Fudan University, \\ 
         \email{eetchen@fudan.edu.cn}
         \and
         $^\ast$This work was completed under the supervision of Dr. Hongyuan Zhu at A*STAR. Chuangguan~Ye and Hongyuan~Zhu are with equal contribution.\\
         $^\dagger$Tao Chen is the corresponding author.\\
}


\date{Received: date / Accepted: date}

\maketitle
    %================ fig: fig1 ================%
    \begin{figure}[htb]
    \centering
    \includegraphics[width=0.45\textwidth]{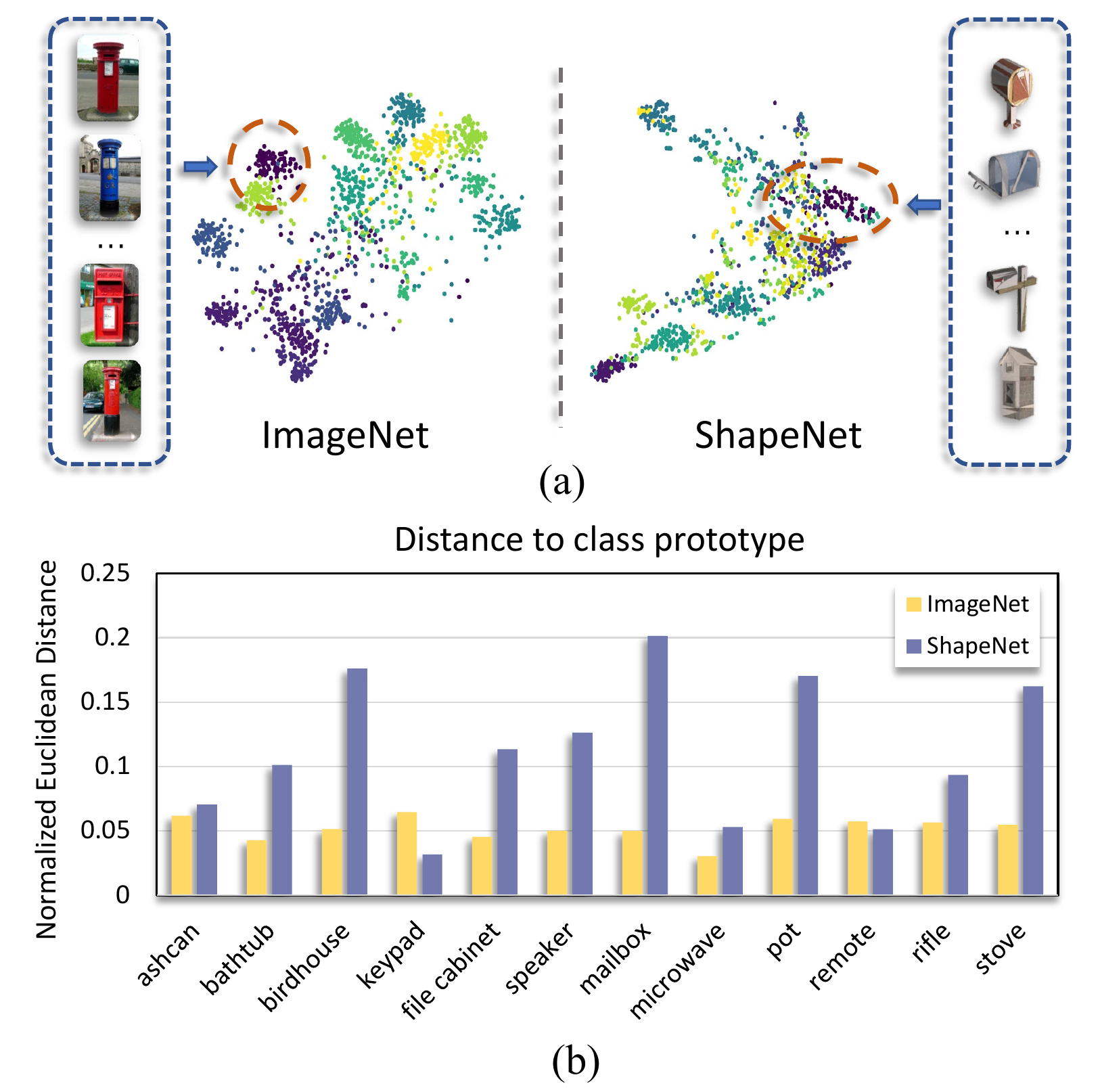} 
    \caption{The challenge of few-shot 3D point cloud classification. (a) is the T-SNE on ImageNet~\cite{russakovsky2015imagenet} and ShapeNet~\cite{chang2015shapenet} using pre-trained 2D image and 3D point cloud features respectively with the “Mailbox” class highlighted. (b) shows the mean normalized Euclidean distance between each example and its class prototype in the visual and point cloud embedding space, respectively. The embedding quality of 2D visual domain is much higher than 3D point cloud domains because image-based pre-trained models, such as ResNet~\cite{he2016deep}, use deeper networks trained on millions of images, whereas point cloud-based models, such as PointNet~\cite{qi2017pointnet}, use shallower networks trained on only a few hundred point clouds with subtle inter-class differences and high intra-class variations~\cite{cheraghian2020transductive}.}\label{fig:challenge}   \vspace{-20pt}
    \end{figure} 
    %================ fig: fig1 ================%
    
\begin{abstract}
    In recent years, research on few-shot learning (FSL) has been fast-growing in the 2D image domain due to the less requirement for labeled training data and greater generalization for novel classes. However, its application in 3D point cloud data is relatively under-explored. Not only need to distinguish unseen classes as in the 2D domain, 3D FSL is more challenging in terms of irregular structures, subtle inter-class differences, and high intra-class variances {when trained on a low number of data.} Moreover, different architectures and learning algorithms make it difficult to study the effectiveness of existing 2D FSL algorithms when migrating to the 3D domain. In this work, for the first time, we perform systematic and extensive investigations of directly applying recent 2D FSL works to 3D point cloud related backbone networks and thus suggest a strong learning baseline for few-shot 3D point cloud classification. Furthermore, we propose a new network, Point-cloud Correlation Interaction (PCIA), with three novel plug-and-play components called Salient-Part Fusion (SPF) module, Self-Channel Interaction Plus (SCI+) module, and Cross-Instance Fusion Plus (CIF+) module to obtain more representative embeddings and improve the feature distinction. These modules can be inserted into most FSL algorithms with minor changes and significantly improve the performance. Experimental results on three benchmark datasets, ModelNet40-FS, ShapeNet70-FS, and ScanObjectNN-FS, demonstrate that our method achieves state-of-the-art performance for the 3D FSL task. Code and datasets are available at  \url{https://github.com/cgye96/A\_Closer\_Look\_At\_3DFSL}.

\keywords{Machine Learning \and Few-Shot Learning \and Meta Learning \and Point Cloud Classification}
\end{abstract}
%%%%%%%%%%%%%%%%%%%%%%%%%%%%%%%%%%%%%%%%%%%%%%%%%%%%%%%%%%%%%%%%%%%%%%%%%

%%%%%%%%%   Introduction  %%%%%%%%%
\section{Introduction}
\label{sec:intro}
    Deep learning models have achieved promising performance on various computer vision (CV) and natural language processing (NLP) tasks, with the advent of powerful computing  resources and large-scale annotated datasets. The 3D point cloud understanding studies with deep learning techniques are also fast-growing~\cite{qi2017pointnet,qi2017pointnet++,li2018pointcnn,liu2019relation,zhang2022riconv++,cosmo20223d}. Unlike traditional point cloud recognition algorithms~\cite{johnson1999using,zhong2009intrinsic,rusu2009fast} that {extract shape features based on the hand-crafted operators, deep-learning-based methods can get more informative descriptors for point cloud instances from} shape projections~\cite{yu2018multi,qi2016volumetric,feng2018gvcnn} or raw points~\cite{qi2017pointnet,qi2017pointnet++,liu2019relation,li2018pointcnn,wang2019dynamic} with deep networks, which achieve better performance.
    
    However, {deep-learning-based methods have two crucial issues} for point cloud classification. Firstly, {the deep-learning technique usually requires large-scale labeled data to train}, but it is cumbersome and costly to annotate extensive point cloud data.
    Secondly, the models trained on base classes often fail to generalize to novel or unseen classes.
    To overcome the data annotation problem, data-augmentation techniques~\cite{tanner1987calculation}, semi-supervised learning~\cite{chapelle2009semi}, or other learning paradigms~\cite{liao2021point} have been introduced to reduce laborious annotation by efficiently augmenting labeled examples or reasonably utilizing unlabeled examples. Point-augment~\cite{li2020pointaugment} is  proposed by leveraging an auto-augmentation framework to address the point cloud classification problem, which formulates a learnable point augmentation function  {using} a shape-wise transformation and a point-wise displacement. PointMixup~\cite{chen2020pointmixup} is to generate new point cloud examples by assigning optimal paths function to mix up two objects. However, they may not obtain promising performances in unseen or novel classes.
     
    On the other hand, we humans can distinguish novel classes with extremely few labeled examples, which inspires the research in few-shot learning (FSL)~\cite{snell2017prototypical,sung2018learning,garcia2017few,vinyals2016matching,SachinRavi2017OptimizationAA,finn2017model,lee2019meta,li2020transferrable,ye2021learning,xu2022attribute}. The FSL generalizes deep networks to novel classes with limited annotated data, which has achieved great success in the 2D image domain. Generally, most current 2D FSL algorithms adopt metric-based~\cite{snell2017prototypical,sung2018learning,garcia2017few,chen2021hierarchical} or optimization-based ~\cite{SachinRavi2017OptimizationAA,finn2017model,lee2019meta} frameworks to learn transferable knowledge and propagate them to new tasks, e.g., Relation Net~\cite{sung2018learning}introduces a learnable non-linear distance metric model to compare query and support images, and MAML~\cite{finn2017model} is to learn a model-agnostic parameter initialization for quickly fine-tuning with few gradient update steps. 
    Moreover, the generalized few-shot learning framework (GFSL)~\cite{ye2021learning} and any-shot learning~\cite{xu2022attribute} are recently introduced to improve the  model's generalization ability to novel classes with few training images. However, compared to the success of FSL in the 2D image domain, its study in 3D data is still relatively under-explored with the following two challenges.
     
    First, different from 2D images defined in structured grids with rich texture and appearance information, {a point cloud instance is a group of irregular, unordered, and unstructured points in 3D Euclidean space}~\cite{guo2020deep}. Such characteristics make it challenging  to identify the critical local areas in a point cloud object, increasing the difficulty of extracting the representative and robust object features. Further, when the 3D point cloud meets with few-shot scenarios, to what extent existing point cloud architecture and learning algorithms can perform for 3D FSL is still unknown. 
    
    Second, most 2D FSL algorithms can {generate more distinguishing embeddings by pre-training deeper networks on large-scale based class data}~\cite{chen2019closer}, e.g., tieredImageNet~\cite{ren2018meta} contains 608 classes and more than 800,000 examples in total. By contrast, most 3D datasets contain smaller-scale annotated data, e.g., ModelNet40~\cite{wu20153d} only has a total number of 12,311 examples and 40 classes. Hence the point cloud learning networks trained on a low number of data may generate poor-quality 3D feature clusters, with subtle inter-class differences and high intra-class variations, as shown in Fig.~\ref{fig:challenge}. Therefore, how to address these issues requires further exploration. 
    
    In this work, for the first time,  we systematically study 3D FSL by performing extensive comparison experiments of certain typical 2D FSL algorithms on different kinds of point cloud learning architectures, and provide comprehensive benchmarks and a strong baseline for few-shot point cloud classification. Moreover, we propose a novel method, Point-cloud Correlation Interaction (PCIA), with three plug-and-play components called the Salient-Part Fusion (SPF) module, Self-Channel Interaction Plus (SCI+) module, and Cross-Instance Fusion Plus (CIF+) module to address the issues of subtle inter-class differences and high intra-class variance causing by the limited number of training data. Concretely, the SPF module is designed to generate more representative embeddings for unstructured point cloud instances by fusing local salient-part features. The SCI+ module and CIF+ module can further improve the feature distinction by exploring the self-channel and cross-instances correlations. Extensive experiments on the three newly proposed 3D FSL benchmark datasets, including ModelNet40-FS, ShapeNet70-FS, and ScanObjectNN-FS, show that our proposed modules can be inserted into certain FSL algorithms flexibly with minor changes and significantly improve their performance.
    
    Our key contributions in this work can be briefly summarized as follows:
    \vspace{-2pt}
    \begin{itemize}
        \item We provide systematic and comprehensive benchmarks and studies for 3D FSL, in terms of the different FSL learning frameworks and point cloud network architectures. We suggest a strong baseline for the few-shot 3D point cloud classification task. 
        \item To address the challenges of 3D FSL, we propose three plug-and-play components, including the Salient-Part Fusion (SPF) module, Self Channel Interaction Plus (SCI+) module, and Cross Instance Fusion Plus (CIF+) module, which can be inserted into certain FSL algorithms with significant performance improvement.
        \item Our proposed network achieves state-of-the-art performance on the three newly split benchmark datasets of 3D FSL, including two synthetic datasets, ModelNet40-FS and ShapeNet70-FS, and a real-world scanning dataset ScanobjectNN-FS, which improves the performance by about 10\% for 1-shot and 4\% for 5-shot. 
    \end{itemize}
    
    Note that this paper is an extended version of our recent conference paper~\cite{ye2022makes}. The significant extensions include \tb{(1)}  constructing an additional benchmark for few-shot 3D point cloud classification on a real-world scanning dataset ScanobjectNN-FS, and adding several recent relevant SOTA methods as comparison baselines (e.g., Point-BERT~\cite{yu2021point});  \tb{(2)} designing a new Salient-Part Fusion (SPF) module to capture the fine details of a point cloud object, by fusing local salient-part features (see Section~\ref{subsec:SPF_module}); \tb{(3)} proposing a new Self Channel Interaction Plus (SCI+) module to mine the task-related information, and use it to enhance the inter-class difference by channel interaction. (see Section~\ref{subsec:SCI+_module});\tb{(4)} extending the Cross-Instance Fusion (CIF) module with a new branch further exploring the correlation among different instances, to become a new Cross-Instance Fusion Plus (CIF+) module. (see Section~\ref{subsec:CIF_module});
    \tb{(5)} providing more extensive experiments (see Section~\ref{subsec:Experimental_results}) and deep analyses (see Section~\ref{subsec:Ablative_Analysis}, ~\ref{subsec:Insight_Analysis} and ~\ref{subsec:Visualization_Analysis}) on model performance, complexity, robustness, etc.
    
    {We organize the rest of the paper as follows.} Section~\ref{sec:liter} formulates the problem definition of 3D FSL and reviews the related works for 3D point cloud classification and 2D FSL. {Section~\ref{sec:empirical_study} performs systematic benchmark to investigate the performance of recent state-of-the-art 2D FSL algorithms on 3D data with different kinds of point cloud learning backbones.} Section~\ref{sec:approach} is to introduce our pipeline for 3D FSL and more details about the proposed method. Extensive experiments and discussions are presented in Section~\ref{sec:experiment}. Some concluding remarks are finally drawn in Section~\ref{sec:conclusion}.

%%%%%%%%%   Problem Definition and Related Work  %%%%%%%%%  
\section{Problem Definition and Related Work}
\label{sec:liter}
\subsection{Problem Definition}
\label{sec:problemdefinition}
    
    We first define a point cloud instance $x$ with its label $y$ as $(x, y)$, where $x \in \mathbb{R}^{n\times 3}$ is an irregular point set containing $n$ points represented with $xyz$ coordinates.
    In the customary $N$-way $K$-shot $Q$-query few-shot learning setting~\cite{chen2019closer}, the aim of FSL algorithms is to meta-train a predictor which can be generalized to new unlabeled query examples by few labeled support examples. We denote the labeled support examples as support set $\mathcal{S} = \{(x_i,y_i)\}_{i=1}^{N_s= N \times K}$, containing $N$ classes with $K$ examples for each class, and denote the unlabeled query examples as the query set $\mathcal{Q}  = \{(x_i,y_i)\}_{i=1}^{N_q=N \times Q}$, containing the same $N$ classes with $Q$ examples for each class.
    
    For few-shot point cloud classification, we adopt the meta-learning paradigm with a set of meta-training episodes $\mathcal{T}=\{(\mathcal{S}_{i}, \mathcal{Q}_{i})\}_{i=1}^{I}$ by optimizing following objectives:
    \begin{ceqn}
    \begin{align}
    \label{eq:optimization}
	    \theta^*,\phi^* =\underset{\theta ,\phi}{\arg\min}\mathcal{L}\left(\mathcal{T};\theta ,\phi \right),
    \end{align}
    \end{ceqn}
    where $\mathcal{T}$ are sampled from the training set and $\mathcal{L}$ denotes the cross-entropy loss function defined as:
    \begin{ceqn}
    \begin{align}
    \label{eq:cross_entropy}
	    \mathcal{L}\left( \mathcal{T};\theta ,\phi \right) =\mathbb{E}_\mathcal{T}\left[ -\log p\left( \hat{y}=c|x \right) \right],
	\end{align}
    \end{ceqn}
    with the prediction  $p\left( \hat{y}=c|x \right)$ can be given by:
    \begin{ceqn}
    \begin{align}
    \label{eq:prediction}
        p\left( \hat{y}=c|x \right) =softmax\left( \mathcal{C}_{\theta}\left( \mathcal{F}_{\phi} \left( x \right) \right) \right), 
    \end{align}
    \end{ceqn}
    where $x$ is the input point cloud instance, $\hat{y}$ is the predicted label and $c$ is the ground truth label. The $\mathcal{F}$ is the backbone network parameterized by $\phi$ and $\mathcal{C}$ is the classifier parameterized by $\theta$.
    
    Once meta-training is finished, generalization of the predictor is evaluated on meta-testing episodes $\mathcal{V}=\{( \mathcal{S}_{j},\mathcal{Q}_{j})\}_{j=1}^{J}$, which are sampled from the testing set. Note that we denote the classes in $\mathcal{T}$ as base classes and the classes in $\mathcal{V}$ as novel classes. The base classes are disjoint with the novel classes.

    Therefore, there are two crucial challenges in Point Cloud FSL:
    1) how to properly represent the point cloud data for few-shot learning; 2) how to effectively transfer the knowledge gained in meta-training episodes to meta-testing episodes, in the case of subtle inter-class differences and high intra-class variance. In the following section, we will review the recent efforts in 3D point cloud classification and 2D few-shot learning.

\subsection{Related Work}
\label{sec:relatedwork}
    \subsubsection{3D Point Cloud Classification}
    \vspace{-3pt}
    Unlike the traditional point cloud recognition methods~\cite{johnson1999using,zhong2009intrinsic}, extracting features based on the hand-crafted operators, deep-learning-based algorithms can get more informative descriptors for point cloud instances with deep networks, which has attracted much attention.  Many studies have made significant progress in adopting deep networks for point cloud recognition. According to the fashion of feature extracting from structured data or unordered points, existing deep-learning-based algorithms can be divided into two major categories, the projection-based methods and the point-based methods.
    
    The projection-based methods first convert the irregular points into  structured representations, such as multi-view images~\cite{su2015multi,yu2018multi,chen2018veram}, voxel~\cite{maturana2015voxnet,riegler2017octnet}, hash tables~\cite{shao2018h} or lattices~\cite{su2018splatnet,rao2019spherical}, and then extract view-wise or structural features with the typical 2D or 3D CNN networks.
    VERAM~\cite{chen2018veram} presents a view-enhanced recurrent attention model to select a sequence of views for 3D classification. OctNet~\cite{riegler2017octnet} designs a hybrid grid-octree to represent point cloud objects, and H-CNN~\cite{shao2018h} uses hierarchical hash tables and convolutional neural networks to facilitate shape analysis.
    However, these approaches may encounter the issue of explicit information loss during the conversion phase or higher memory consumption during the feature learning phase~\cite{guo2020deep}.
    
    The point-based methods, in contrast, directly take the irregular points as input and embed the point cloud object by exploring the point-wise relationship with deep networks. PointNet~\cite{qi2017pointnet} is the first model to extract point cloud features using a deep network on raw points. Furthermore, the following works, such as PointNet++~\cite{qi2017pointnet++}, PointCNN~\cite{li2018pointcnn},  RSCNN~\cite{liu2019relation}, DensePoint~\cite{liu2019densepoint}, and DGCNN~\cite{wang2019dynamic}, further capture the local structural or geometric information using convolution-base networks or graph-based networks for learning more representative features.
    Moreover, RIConv++~\cite{zhang2022riconv++} introduces a Rotation-Invariant Convolution by modeling the internal relationship between the selected points and their local neighbors to improve feature descriptions.
    However, these deep-learning-based methods require large-scale labeled data to train. The recognition ability of deep networks depends on the contents of the training data, and hence they may have a poor generalization of novel classes never seen before.

    Furthermore, recent works~\cite{sharma2020self,stojanov2021using,yu2021point} attempt to improve the learning ability of deep networks in 3D point cloud learning with few labeled training data.\cite{sharma2020self} introduces a self-supervised pre-training method to obtain point-wise features for few-shot point cloud learning. LSSB~\cite{stojanov2021using} takes SimpleShot~\cite{wang2019simpleshot} as the baseline, and improves the low-shot image classification generalization performance by learning a discriminative embedding space with 3D object shape bias. Point-BERT~\cite{yu2021point} proposes to generalize the Transformers~\cite{self-attention} architecture to 3D point cloud, and introduces a BERT-style~\cite{devlin2018bert} pre-training technique to pre-train a Transformers-based network with a Mask-Point-Modeling (MPM) task, achieving the SOTA performance on several downstream tasks.
    \cite{HengxinFeng2022EnrichFF} is to propose a network for few-shot point cloud classification by designing a feature supplement module to enrich the geometric information and using a channel-wise attention module to aggregate multi-scale features. However, they calculate the channel-wise attention with points' features directly, which may encounter the issue of larger network parameters and higher computational consumption. Besides, the attention mechanism in~\cite{HengxinFeng2022EnrichFF} only considers the point-wise relationship in a single sample and neglects the correlation between different samples. 
    
    Different from these works, we provide comprehensive benchmarks and studies for few-shot point cloud classification, by systematically reviewing certain typical FSL algorithms migrating to different 3D point cloud learning networks. We also propose three effective and light modules for 3D FSL, including the Salient-Part Fusion (SPF) module, the Self-Channel Interaction Plus (SCI+) module and the Cross-Instance Fusion Plus (CIF+) module, which can be easily inserted into most FSL algorithms with significant performance improvement.

    \subsubsection{2D Few-Shot Learning} 
    Recently, research on few-shot learning (FSL) has been fast-growing in the 2D image domain due to the less demand for labeled training data and better generalization for novel classes. Generally, most 2D FSL algorithms adopt a meta-learning strategy to learn the transferable meta-knowledge from base classes to novel classes, roughly divided into metric-based approaches and optimization-based approaches.

    The metric-based approaches try to learn a distinguishing feature space in which examples belonging to the same class are closer, or design efficient metric functions to measure the similarity of features. Matching Net~\cite{vinyals2016matching} designs a bidirectional Long-Shot-Term Memory (LSTM) module to obtain global contextual embeddings and predicts the query examples' label with the Cosine Similarity Distance. Prototypical Net~\cite{snell2017prototypical} introduces the concept of class-prototype, which is the mean of the support features of each class, and uses the Squared Euclidean Distance to classify query examples, demonstrating better performance than Matching Net~\cite{vinyals2016matching}. Moreover, Relation Net~\cite{sung2018learning} and FSLGNN~\cite{garcia2017few} further propose learnable metric modules to obtain relation scores or model the instance-wise relationship of labeled support examples and query examples.
    
    On the other hand, the optimization-based approaches attempt to fast adapt network parameters or optimization strategies to novel tasks with few gradient update steps.
    Meta-leaner~\cite{SachinRavi2017OptimizationAA} replaces the stochastic gradient descent optimizer with an LSTM-based model. 
    MAML~\cite{finn2017model} tries to learn a transferable parameter initialization that can be fast generalized to new tasks with few fine-tuning steps. MetaOptNet~\cite{lee2019meta} meta-trains a feature-relevant Support-Vector Machine (SVM) predictor for query examples by incorporating a differentiable quadratic programming solver.
    
    Recently, many efforts have been devoted to further promoting the 2D FSL. ~\cite{PuneetMangla2019s2m2} proposes the S2M2 to use self-supervision and regularization techniques to learn relevant feature manifolds for FSL. \cite{CarlDoersch2020CrossTransformersSF} proposes the CrossTransformers to design a Transformer based network to find spatially corresponding between query and support samples. \cite{YinboChen2021MetaBaselineES} is to introduce Meta-baseline with a cosine metric classifier with learnable weight, showing better performance than Squared Euclidean Distance. \cite{XuLuo2022ChannelIM} proposes SimpleTrans to use a simple transform function to adjust the weight of different channels to  alleviate the channel bias problem in FSL.
    
    Differently, in this work, we empirically conduct a comprehensive study of different 2D FSL approaches migrating to few-shot point cloud classification tasks, and suggest a strong baseline for 3D FSL validated on three new splits of benchmark datasets.

\vspace{-10pt}

%%%%%%%%%   Systematic Benchmark  %%%%%%%%%  
\section{Systematic Benchmark}
\label{sec:empirical_study}
    This section introduces three new splits of benchmark datasets for 3D FSL and systematically studies certain typical 2D FSL methods on these benchmark datasets with different typical point-based network backbones.  The details of experimental settings and implementation are described in Section~\ref{sec:experiment}.

    \subsection{Benchmark Datasets for 3D FSL}
    \label{sec:Benchmark}

    ModelNet40~\cite{wu20153d} and ShapeNetCore~\cite{chang2015shapenet} are two widely used 3D synthetic model benchmark datasets for general 3D point cloud classification, containing 40 and 55 categories with a total number of 12,311 and 51,192 objects, respectively. ScanObjectNN~\cite{uy2019revisiting} is the first real-world point cloud classification dataset containing 15 categories with 2,902 unique objects in total, which are selected from the indoor datasets (SceneNN~\cite{uy2019revisiting} and ScanNet~\cite{dai2017scannet}).
    
    However, there are two issues in these datasets for 3D FSL: 1) existing splits of the training set and testing set contain overlapping classes, which should be disjunctive in the few-shot setting~\cite{vinyals2016matching}; 2) the number of examples in each class is imbalanced. To meet the FSL setting requirement and evaluate the performance objectively, we introduce new splits of these datasets and construct three benchmark datasets, ModelNet40-FS, ShapeNet70-FS and ScanObjectNN-FS, for 3D few-shot point cloud classification under different scenarios. Statistics of the proposed benchmark datasets are listed in Table~\ref{table:dataset}.
    
    \tb{ModelNet40-FS} includes the same 40 classes proposed in the ModleNet40~\cite{wu20153d}, and we carefully select 30 classes as the base class set for training and the other 10 classes as the novel class set for testing, to ensure that all the base classes are sufficiently distinct from the novel classes.
    
    \tb{ShapeNet70-FS} is a larger dataset for 3D FSL. We first select 48 categories with a sufficient example size (at least 80 examples in a class) from ShapeNetCore~\cite{chang2015shapenet} and then increase the class number to 70 by splitting some broad categories into several fine-grained subcategories according to the taxonomy offered by~\cite{chang2015shapenet}. After that, these are split into 50 base classes for training and 20 novel classes for testing. Note that fine-grained classes separated from the same super-category are all in the training set or testing set, ensuring that the test classes are not similar to those seen during training.
    
    \tb{ScanObjectNN-FS} is a more challenging dataset for 3D FSL because the objects scanned from the real world usually contain cluttered background noise and large perturbations. There are five variants in ScanObjectNN~\cite{uy2019revisiting} with different degrees of perturbation. We choose the most challenging variant, PB\_T50\_RS, to construct the ScanObjectNN-FS, which consists of 15 categories with 14,298 perturbed objects in total. We evenly split the object classes into three non-overlapping subsets, and each split contains 5 classes.
    
     %================ table: dataset ================%
    \begin{table}[tb]\centering
    \renewcommand{\arraystretch}{1.2}
    \caption{\label{table:dataset} Statistics of the new split benchmark datasets for few-shot point cloud classification.  {$S^i$ denotes the $i$ split of ScanObjectNN-FS.}} 
    \resizebox{0.45\textwidth}{!}{
    \begin{tabular}{c|c|ccc}
   \hline
    \multicolumn{2}{c|}{Bnechmark Datasets for 3D FSL}                 
    & Train  & Test  & Total  \\ \hline  
    \multirow{2}{*}{\bf{ModelNet40-FS}} 
    & Classes    & 30     & 10    & 40     \\ 
    & Instances  & 9,204  & 3,104 & 12,308 \\ \hline
    \multirow{3}{*}{\bf{ShapeNet70-FS}} 
    & Categories & 34     & 14    & 48     \\
    & Classes    & 50     & 20    & 70     \\ 
    & Instances  & 21,722 & 8,351 & 30,073 \\\hline 
    \multirow{4}{*}{\bf{ScanObjectNN-FS}}  
    & Classes    & 10     & 5    & 15    \\ 
    & Instances($S^0$)  &  9,509   & 4,789   & 14,298 \\     
    & Instances($S^1$)  &  9,129   & 5,169  & 14,298 \\    
    & Instances($S^2$) &  9,958  & 4,340     & 14,298 \\    
   \hline                     
    \end{tabular}}  \vspace{-10pt}
    \end{table} 
    %================ table: dataset ================%
    
    \subsection{Baselines}
    In terms of the few-shot point cloud classification, we first study six widely used SOTA 2D FSL methods, including three metric-based methods and three optimization-based methods. Here we introduce more adapting details about these FSL algorithms migrating to 3D point cloud data.

    \textbf{Metric-Based Methods}:
    Following the original setting proposed in \tb{Prototypical Net}~\cite{snell2017prototypical}, we first average the support features as class prototype features and then take the Squared Euclidean Distance to classify the query examples. We train the network end-to-end by minimizing the cross-entropy loss. 
    \tb{Relation Net}~\cite{sung2018learning} proposes a learnable relation module to predict the relation score of support-query pairs. Similar to~\cite{sung2018learning}, we design a learnable relation module {consisting of} two convolutional blocks (($1\times1$ Conv with 128 filters, BN, ReLU), ($1\times1$ Conv with one filter, BN, ReLU)) and two Fully-Connection Layers (128, N) (N is the number of classes at a meta-episode) and use the Mean Square Error loss to regress relation score to ground truth.
    \tb{FSLGNN}~\cite{garcia2017few} first introduces a graphical model into 2D FSL, which is easily extended to variants of frameworks. We construct four Edge-Conv layers with the setting introduced in~\cite{garcia2017few} and adopt the Cross-Entropy loss as the optimal objective.
    
    \textbf{Optimization-Based Methods}: 
    \tb{MAML}~\cite{finn2017model} aims to learn a  {potential parameter initialization that can be quickly generalized to novel classes with few} gradient update steps. We adopt two FC layers (256, N) as the classifier and update the gradient twice during the meta-training stage.
    \tb{Meta-learner}~\cite{SachinRavi2017OptimizationAA} proposes to learn an optimizer for novel examples, which takes two LSTM layers as a meta-learner to replace the stochastic gradient descent optimizer. We follow the same settings introduced in~\cite{SachinRavi2017OptimizationAA} and design two LSTM layers for gradient updating.
    \tb{MetaOptNet}~\cite{lee2019meta} tries to generalize to novel categories with a linear classifier, by using implicit differentiation and dual formulation. We use the multi-class SVM presented in~\cite{crammer2001algorithmic} as the classification head.
    
    \vspace{-5pt}
    \subsection{Training and Testing Strategy}
    \label{sec:train_strategy}
    We train and test the network with the standard episode-based FSL setting ~\cite{chen2019closer}. Firstly, we train the network from scratch with 80 epochs. Each epoch contains 400 meta-training episodes and 600 validating episodes, {which are sampled from the training set data and validating set data randomly.} Each episode contains $N$ classes with $K$ labeled support examples and $Q$ query examples for each class, denoted as the $N$-way $K$-shot $Q$-query setting. Once the meta-training is ended, we test the network with 700 episodes randomly sampled from the testing set data with the same $N$-way $K$-shot $Q$-query setting, and report the mean classification results of the 700 episodes with 95\% confidence intervals. During the testing stage, we also fix the random seed to ensure that all the methods can be compared fairly.
   
    Specially, we employ the 5-fold cross-validation training strategy for ModelNet40-FS and ShapeNet70-FS, {where we randomly divided the training data into five even subsets. Each subset is used at a time as validation data to select the best validation model. Then, we test the best-validation models and the last-training models on testing data, and take the better average performance of the five estimations as the final results.} For ScanObjectNN-FS, we perform 3-fold cross-validation by selecting one split as the testing set and taking the remaining splits as the training set. Therefore, there are three different splits of the  ScanObjectNN-FS, denoted as "$S^0$", "$S^1$", and "$S^2$", respectively.

    %================ table: Pointnet+2DFSL ================% 
    \begin{table*}[thb] \centering
    \renewcommand{\arraystretch}{1.5}
    \caption{\label{table:baseline_fsl} Few-shot point cloud classification accuracy (\%) on three newly proposed benchmark datasets with PointNet~\cite{qi2017pointnet} as the backbone network. All the algorithms are tested on 700 meta-testing episodes with 95\% confidence intervals. Bold means the best result and underline means the second best. The $M$ are the metric-based approaches and $O$ are the optimization-based approaches. PN: parameter number. GFLOPs: the number of floating-point operations. TPS: inference tasks per second on one NVIDIA 2080Ti GPU.}
    \setlength{\tabcolsep}{1mm}{
    \begin{footnotesize}
    \scalebox{0.85}{
    \begin{tabular}{c c cc cc cc ccc}
    \hline
    \multicolumn{2}{c }{
    \multirow{2}{*}{Method}} & 
    \multicolumn{2}{c }{ModelNet40-FS} & 
    \multicolumn{2}{c }{ShapeNet70-FS} & 
    \multicolumn{2}{c }{ScanObjectNN-FS} &
    \multicolumn{3}{c}{5way-1shot-15query}   \\ \cline{3-11} 
    \multicolumn{2}{c }{} & 5way-1shot   & 5way-5shot 
    & 5way-1shot   & 5way-5shot     
    & 5way-1shot   & 5way-5shot        
    & PN      & GFLOPs    & TPS  
    \\ \hline 
    \multirow{3}{*}{~~M~~}
    & Prototypical Net~\cite{snell2017prototypical}        
    & \tb{65.31$\pm$0.78}    & \underline{79.04$\pm$0.54} 
    & \tb{65.96$\pm$0.81}    & \tb{78.77$\pm$0.67} 
    & 44.75$\pm$0.31         & \tb{59.81$\pm$0.26}
    & \tb{0.15M}  & \tb{6.16}  & \tb{118.23}\\
    & Relation Net~\cite{sung2018learning}     
    & 64.10$\pm$0.72         & 75.75$\pm$0.57 
    & \underline{65.88$\pm$0.85}    & 76.25$\pm$0.71 
    & \underline{45.32$\pm$0.32}    & 55.43$\pm$0.26
    & \underline{0.28M}   & 6.50      & \underline{96.03} \\
    & FSLGNN~\cite{garcia2017few}           
    & 59.69$\pm$0.73         & 76.06$\pm$0.63 
    & 64.98$\pm$0.84         & 76.14$\pm$0.73 
    & 29.91$\pm$0.39         & 32.77$\pm$0.33
    & 2.23M          & 10.44   & 69.02  \\ \hline
    \multirow{3}{*}{~~O~~} 
    & Meta-learner~\cite{SachinRavi2017OptimizationAA}       
    & 58.69$\pm$0.81         & 76.60$\pm$0.65 
    & 62.64$\pm$0.91         & 73.10$\pm$0.80 
    & 43.79$\pm$0.34         & 58.89$\pm$0.26
    & 0.88M         & 6.28     & 9.14  \\
    & MAML~\cite{finn2017model}             
    & 57.58$\pm$0.89         & 77.95$\pm$0.62 
    & 59.20$\pm$0.88         & 75.10$\pm$0.75 
    & \tb{46.65$\pm$0.37}    & \underline{59.55$\pm$0.23}
    & 0.68M    & \underline{6.19}     & 38.71 \\
    & MetaOptNet~\cite{lee2019meta}       
    & \underline{64.99$\pm$0.87}    & \tb{79.54$\pm$0.61} 
    & 65.08$\pm$0.89         & \underline{77.81$\pm$0.75} 
    & 45.02$\pm$0.31         & 59.39$\pm$0.24
    & \tb{0.15M} & \tb{6.16}   & 11.95 \\ 
    \hline 
    \end{tabular}}
    \end{footnotesize}}  \vspace{-5pt}
    \end{table*}
    %================ table: Pointnet+2DFSL ================%

    %================ table: protoNet+3Dbackbone ================% 
    \begin{table}[tb] \centering
    \renewcommand{\arraystretch}{1.5}
    \caption{\label{table:baseline_backbone} Comparison results (accuracy \%) of vanilla few-shot classification (noted as $\divideontimes$ ) and Prototypical Net~\cite{snell2017prototypical} (noted as $\ddagger$ ) on three newly proposed benchmark datasets with different backbones.} 
    \setlength{\tabcolsep}{1mm}{
    \begin{footnotesize}
    \scalebox{0.8}{
    \begin{tabular}{ c cc  cc cc}
    \hline 
    \multicolumn{1}{c }{
    \multirow{2}{*}{Method}}  & 
    \multicolumn{2}{c}{ModelNet40-FS} & \multicolumn{2}{c}{ShapeNet70-FS} &
    \multicolumn{2}{c}{ScanObjectNN-FS}
    \\ \cline{2-7}   
    & 5w-1s   & 5w-5s
    & 5w-1s   & 5w-5s   
    & 5w-1s   & 5w-5s   
    \\ \hline  
     {$^\divideontimes$PointNet~\cite{qi2017pointnet}}  
    & {60.03}	& {78.25}	& {60.53}	& {77.91}	& {40.00}	& {56.83} \\
    $^\ddagger$PointNet~\cite{qi2017pointnet}  
    &\tb{65.31}  &\tb{79.04}  &\tb{65.96}  &\tb{78.77}  &\tb{44.75}  &\tb{59.81} \\ \hline 
    
     {$^\divideontimes$PointNet++~\cite{qi2017pointnet++}}
    & {61.18}	& {82.31}	& {50.85}	& {79.10}	& {34.30}	& {57.74} \\
     $^\ddagger$PointNet++~\cite{qi2017pointnet++}
    &\tb{64.96}   &\tb{83.66}  &\tb{66.33} &\tb{80.95}  
    & \tb{46.83}  & \tb{62.27} \\\hline 
    
    {$^\divideontimes$PointCNN~\cite{li2018pointcnn}}   
    & {59.28}	& {74.77}	& {48.94}	& {75.96}	& {29.78}	& {56.68} \\
    $^\ddagger$PointCNN~\cite{li2018pointcnn}   
    &\tb{60.38}       &\tb{76.95}  
    &\tb{64.02}       &\tb{76.34}  
    &\tb{46.09}       &\tb{59.11} \\\hline 
    
     {$^\divideontimes$RSCNN~\cite{liu2019relation}}      
   & {61.29}	& {80.06}	& {57.37} 	& {77.96}	& {31.88}	& {53.81} \\
    $^\ddagger$RSCNN~\cite{liu2019relation}      
    & \tb{69.72}  &\tb{84.79}
    & \tb{68.66}  & \tb{82.55} 
    & \tb{48.58}  &\tb{63.89}  \\\hline 
    
    {$^\divideontimes$DensePoint~\cite{liu2019densepoint}} 
   & {57.30}	& {78.06}	& {58.42}	& {79.43}	& {30.10}	& {49.91} \\ 
   $^\ddagger$DensePoint~\cite{liu2019densepoint} 
    &\tb{66.99}       &\tb{82.85}  
    &\tb{65.81}       &\tb{80.74}   
    &\tb{43.68}       &\tb{57.36} \\ \hline 
    {$^\divideontimes$DGCNN~\cite{wang2019dynamic}}      
    & {63.52}	& {84.00}	& {61.57}	& {81.55}	& {42.46}	& {60.98} \\ 
    $^\ddagger$DGCNN~\cite{wang2019dynamic}      
    &  \tb{69.95}  &  \tb{85.51} 
    &  \tb{69.03}  & \tb{82.08} 
    &\tb{46.28}       &  \tb{63.94} \\ 
    \hline 
    \end{tabular}}
    \end{footnotesize}} \vspace{-10pt}
    \end{table}
    %================ table: protoNet+3Dbackbone ================%
    
    \vspace{-5pt}
    \subsection{State-of-the-Art 2D FSL on Point Cloud}
    Firstly, we investigate the performance of directly applying widely used SOTA 2D FSL algorithms to the few-shot point cloud classification task on our newly split benchmark datasets. Given its simplicity and efficiency, we adopt PointNet~\cite{qi2017pointnet} as the embedding backbone network, and divide the state-of-the-arts into the following groups:

    \begin{itemize}
        \item Metric-based approaches \textbf{M}:  Prototypical Net~\cite{snell2017prototypical}, Relation Net~\cite{sung2018learning}, FSLGNN~\cite{garcia2017few}
        \item Optimization-based approaches \textbf{O}: Meta-learner~\cite{SachinRavi2017OptimizationAA},  MAML~\cite{finn2017model}, MetaOptNet~\cite{lee2019meta}
    \end{itemize}

    We conduct the experiments under a canonical 5-way setting presented in Section~\ref{subsec:setting}. The comparison results of mean accuracy and complexity, reported in Table~\ref{table:baseline_fsl},
    show that the metric-based approaches outperform the optimization-based approaches in most few-shot classification settings on 3D point cloud data.

    The reason may be that the optimization-based approaches are more sensitive to the architecture of neural networks and need laborious hyper-parameter adjusting to get good generalization, as discussed in recent research in 2D image domains \cite{antoniou2018train}. The optimization-based methods also have lower inference speeds because they need to update the network's parameters in the inference stage.

    Moreover, Prototypical Net~\cite{snell2017prototypical} with PointNet~\cite{qi2017pointnet} as backbone can achieve the top performance of $65.31\%$, $65.96\%$ and $59.81\%$ respectively for all datasets, with less computational consumption and high inference speed. Noted that Prototypical Net~\cite{snell2017prototypical} and MetaOptNet~\cite{lee2019meta} have the same parameter size, as they share the same backbone network and use the parameter-free Square Euclidean Distance and Support Vector Machine (SVM) as the meta classifier, respectively. However, solving the quadratic programming of SVM in MetaOptNet~\cite{lee2019meta} is very computationally expensive and time-consuming.

    In summary, Prototypical Net~\cite{snell2017prototypical} has a better trade-off between performance and complexity, but there is still a large room for further improvement. 
    
    \subsection{Influence of Backbone Architecture on FSL}
    We further analyze the influence of different point cloud backbones on 3D FSL by comparing the vanilla few-shot classification to  Prototypical Net~\cite{snell2017prototypical}. We select three types of state-of-the-art 3D networks including:
    \begin{itemize}
        \item Pointwise-based: PointNet~\cite{qi2017pointnet} and PointNet++~\cite{qi2017pointnet++}.
        \item Convolution-based: PointCNN~\cite{li2018pointcnn}, RSCNN~\cite{liu2019relation} and DensePoint~\cite{liu2019densepoint}.
        \item Graph-based: DGCNN~\cite{wang2019dynamic}.
    \end{itemize}

    For vanilla few-shot classification, we first pre-train these classification networks (including their backbone and the classifier head) on base class data without any architecture change, and then evaluate the classification results on novel class data by fine-tuning the classifier head with labeled support samples in each meta-testing episode. Specifically, We pre-train the networks with their default parameter settings for 100 epochs on each dataset and fine-tune the classifier head for 40 epochs.
    
    For Prototypical Net~\cite{snell2017prototypical}, we remove the last classifier layers of these networks and train the backbone from scratch with the meta-learning paradigm introduced in Sec.~\ref{sec:train_strategy}. Then, embeddings extracted by these backbones are fed into the Prototypical Net~\cite{snell2017prototypical} for the few-shot classification.
    
    Table~\ref{table:baseline_backbone} reports the comparison results of vanilla few-shot classification (noted as $\divideontimes$) and the prototypical Net~\cite{snell2017prototypical} (noted as $\ddagger$). One can clearly obverse that the DGCNN~\cite{wang2019dynamic} outperforms other backbone networks in the vanilla classification setting, illustrating that the DGCNN~\cite{wang2019dynamic} has good potential in feature extraction and generalization. Moreover, the graph-based network DGCNN~\cite{wang2019dynamic} also achieves better performance under most settings in the Prototypical Net~\cite{snell2017prototypical}. The reason can be attributed to the dynamic update of the point-wise connection graph in the feature space and the efficient feature extracting in the EdgeConv layers. Thus the DGCNN~\cite{wang2019dynamic} can obtain more informative features compared with other backbone networks.
    
    Besides, the prototypical Net~\cite{snell2017prototypical} is found to significantly improve the performance in some settings and provide higher performance gains for weaker backbones, for example, 17\% for RSCNN~\cite{liu2019relation} and 16\% for PointCNN~\cite{li2018pointcnn} under the 5w-1s setting on ScanObjectNN-FS dataset. Therefore, we use Prototypical Net~\cite{snell2017prototypical} with DGCNN~\cite{wang2019dynamic} as the strong baseline for 3D FSL.
    
    %================ fig: pipeline ================%
    \begin{figure*}[htp]
    \begin{center}
    \includegraphics[width=0.9\linewidth]{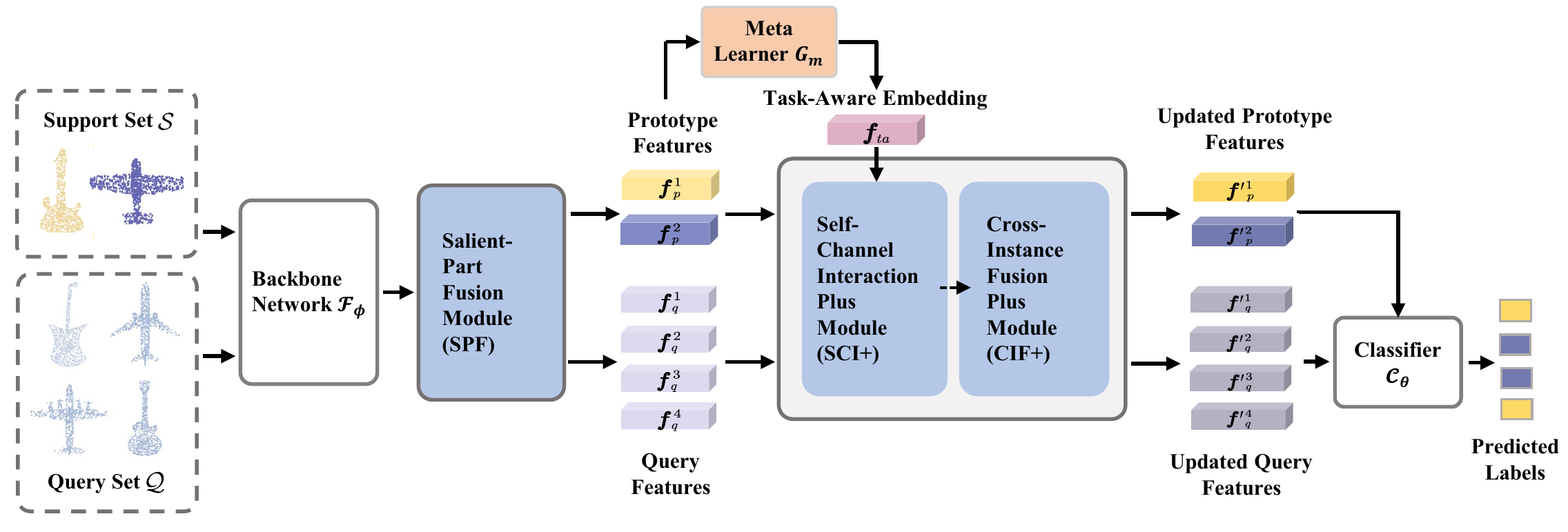}   \vspace{-10pt}
    \end{center}
    \caption{The schematic of our proposed PCIA framework for 3D few-shot point cloud classification, which consists of three main components for addressing the challenges in 3D FSL: Salient-Part Fusion (SPF) module, Self-Channel Interaction Plus (SCI+) module and Cross-Instance Fusion Plus (CIF+) module. For clarity, only the 2-way 1-shot 2-query setting is presented here.}
    \label{fig:pipeline} 
    \end{figure*}
    %================ fig: pipeline ================%

%%%%%%%%%   Approach  %%%%%%%%%   
\section{Approach}
\label{sec:approach}

    \subsection{Overview}
    \label{sec:pipeline}
\begin{figure*}[htp]
    \begin{center}
    \includegraphics[width=0.7\linewidth]{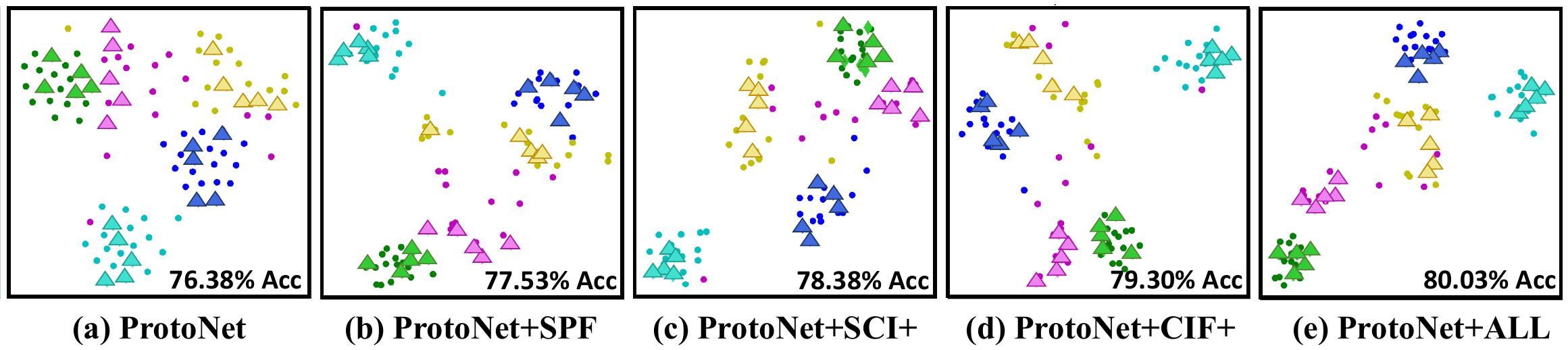}
    \end{center}    \vspace{-10pt}
    \caption{The t-SNE comparisons of feature distribution on the same episode after using different modules. $\triangle$ stands for the support features, $\bullet$ represents the query features. We use the same color to represent the same class. }
    \label{fig:tsne}   \vspace{-10pt}
    \end{figure*}

    Given the baselines in the previous section, 3D FSL still faces certain challenges: 1) most existing models are developed under a lab-controlled assumption, where point clouds are uniformly sampled from the surface of synthetic CAD models without any perturbation. Actually, point clouds collected from the real-world environment usually are incomplete and contain cluttered background points; 2) clusters of 3D features tend to have subtle inter-class differences and strong intra-class variations when the amount of training data is low; 3) the aforementioned methods independently extract features from support set and query set, without considering the correlations between these two sets. Therefore they may suffer from a huge distribution shift, as demonstrated in Fig. \ref{fig:tsne} (a). To address these issues, we create a new and stronger network, Point-cloud Correlation Interaction (PCIA), for 3D FSL classification, as illustrated in Fig. \ref{fig:pipeline}.
    
    %================ fig: t-sne ================%
    %================ fig: t-sne ================%
    
    For the first challenge, we propose a novel salient part fusion (SPF) module to obtain an informative global shape descriptor for unstructured point cloud instances. Specifically, the backbone network $\mathcal{F}_{\phi}$ takes support set $\mathcal{S}$ and query set $\mathcal{Q}$ as input, {and embeds each instance} $x\in \mathbb{R}^{n\times 3}$ into a $d$-dimension point-wise feature map $\boldsymbol{F}_{pw}=\mathcal{F}_{\phi}\left(x \right) = \{\boldsymbol{f}_{pw}^i\} \in \mathbb{R}^{n \times d} $, where $i \in [1, n]$. The feature map $\boldsymbol{F}_{pw}$ are then fed into the proposed SPF module in Section  \ref{subsec:SPF_module}, to fuse local salient features to obtain an informative global shape descriptor $\boldsymbol{f}_g=SPF\left(\boldsymbol{F}_{pw}\right) \in \mathbb{R}^{1\times d}$. After that, we define the prototype feature for class $c_i$ in support set $\mathcal{S}$ as$\boldsymbol{f}_p^i=\frac{1}{|K|}\varSigma _{x_s^k\in \mathcal{S}^{c_i}}SPF\left(\mathcal{F}_{\phi}\left( x_s^k \right) \right)\in \mathbb{R}^{1\times d}$ as the average of its $K$ support examples $\mathcal{S}^{c_i}$, where $\mathcal{S}^{c_i}= \{x_s^1, x_s^2, ..., x_s^K\}$, and the query feature for a query example {$x_q^j$} as $\boldsymbol{f}_q^j=SPF\left(\mathcal{F}_{\phi}\left( x_q^j \right)\right)\in \mathbb{R}^{1\times d}$, where $i\in[1,N]$ and $j\in[1,N_q]$.
    
    However, the first step only separately extracts features from the support set and query set, yet ignores mitigating the feature distribution gap by mining the correlation between these two sets. To address these challenges, our recent work~\cite{ye2022makes} proposed two efficient components called Self-Channel Interaction (SCI) module and Cross-Instance Fusion (CIF) module {to update the prototype feature ${\boldsymbol{f}_p^{i}}$ and query feature ${\boldsymbol{f}_q^{j}}$ by considering the channel-wise and instance-wise  correlation}, which can be easily inserted into different baselines with significant performance improvement. However, they ignore the local task-related information offered by the labeled support examples and the implicit spatial correlation among the input instances.

    Therefore, we newly design a Self-Channel Interaction Plus (SCI+) module in Section \ref{subsec:SCI+_module} and a Cross-Instance Fusion Plus (CIF+) module in Section \ref{subsec:CIF_module} to further refine the prototype feature ${\boldsymbol{f}_p^{i}}$ and query feature by considering the shared and complementary information among prototypes, after which more diverse prototype features $\boldsymbol{f'}_{p}^{i}$ and query features $\boldsymbol{f'}_{q}^{j}$ are generated to improve the distribution of features for better classification, as shown in Fig. \ref{fig:tsne} (e).

    After that, we take Square Euclidean Distance metric function as classifier $\mathcal{C}_{\theta}$, and the probability of predicting label $\hat{y_j}$ for  $\boldsymbol{f'}_q^j$ as class $c_i$ is denoted as:
    \begin{ceqn}
    \begin{align}
    \label{eq:overview_prediction}
        p\left( \hat{y_j}=c_i|\boldsymbol{f'}_q^j \right) =\frac{\exp \left( -d\left( \boldsymbol{f'}_q^j,\boldsymbol{f'}_p^{ {i}} \right) \right)}{\sum_{i=1}^{N}{\exp \left( -d\left( \boldsymbol{f'}_q^j,\boldsymbol{f'}_p^i \right) \right)}},
    \end{align}
    \end{ceqn}
    where $d\left( .,. \right)$ is the Square Euclidean Distance, $\boldsymbol{f'}_p^i$ and  $\boldsymbol{f'}_q^j$ are the updated features generated by the SCI+ and CIF+ module. 
    
    In the end, we can get the cross-entropy loss with Eq.~\ref{eq:cross_entropy} and optimize the network end-to-end by minimizing the following equation:
    \begin{ceqn} \begin{align}
    \label{eq:overview_loss}
        L_{CE} = -\frac{1}{N}\frac{1}{N_q}\sum_i^{N}{\sum_j^{N_q}{\mathbbm{1} \left[ y_j=c_i \right] \log \left( p\left( \hat{y}_j=c_i|\boldsymbol{f'}_q^j \right) \right)}},
    \end{align} \end{ceqn}
    where $N$ and $N_q$  are the numbers of class prototypes and query examples, respectively, $y_j$ is the ground truth of $\boldsymbol{f'}_q^j$, and $\mathbbm{1}$ denotes the Kronecker delta function.
    
    %================ fig: SPF ================%
    \begin{figure*}[t]
    \begin{center}
    \includegraphics[width=0.9\linewidth]{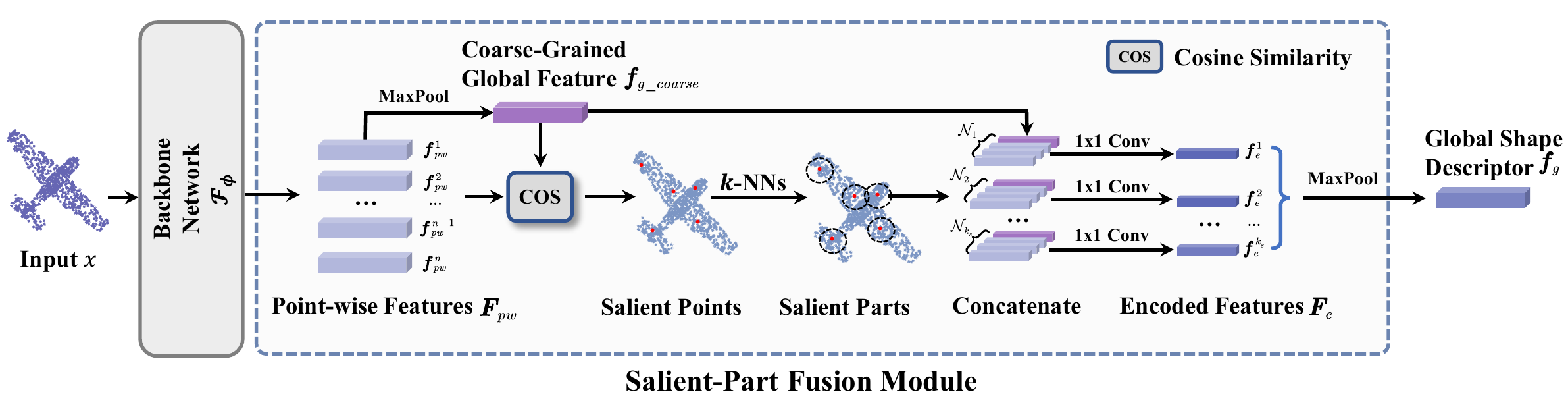}  
    \end{center}  \vspace{-10pt} 
    \caption{An illustration of the Salient-Part Fusion (SPF) module. We first calculate a salient score for each point's feature $\boldsymbol{f}_{pw}^{i}$, $i\in[1,n]$ between the coarse-grained global descriptor { $\boldsymbol{f}_{g\_coarse}$} by using cosine similarity. {Then we select $k_s$ points with the highest salient score and aggregate their $k$-nearest neighbors as the local salient part features {$\mathcal{N}_j$, $j\in[1,k_s]$}.} After that, we concatenate the coarse-grained global feature to each local salient part feature, and adopt a $1\times1$ convolutional layer $g$ to encode the concatenated features as $\boldsymbol{F}_{e}$. Finally, we perform a channel-wise Max-pooling function on $\boldsymbol{F}_{e}$ and obtain the refined global shape descriptor { $\boldsymbol{f}_g$}.   } 
    \label{fig:spf_module} \vspace{-5pt}
    \end{figure*}
    %================ fig: SPF ================% 

    \subsection{Salient-Part Fusion (SPF) Module}
    \label{subsec:SPF_module}
    Since point cloud instances scanned from real-world usually contain cluttered backgrounds and large perturbations, we propose the Salient-Part Fusion (SPF) module to capture the fine details, and learn a more informative shape descriptor for each input point cloud instance by fusing the local salient part features with its coarse-grained global shape feature, as illustrated in Fig. \ref{fig:spf_module}. 
    
    Considering that points from different parts contribute differently to shape description, we tend to select relevant and salient parts to enhance the discrimination of shape descriptor embedding. To this end, we first define the coarse-grained global feature $\boldsymbol{f}_{g-coarse} \in \mathbb{R}^{1\times d}$ for the point cloud $x$, by a channel-wise Max-pooling function performed on $\boldsymbol{F}_{pw}$:
    \begin{ceqn} 
    \begin{align}
    \label{spf_global_feature}
        \boldsymbol{f}_{g\_coarse}=Maxpooling\left(\boldsymbol{F}_{pw}\right) \in \mathbb{R}^{1\times d},
    \end{align} \end{ceqn}
    
    Then, we further define a salient score $a_i$ to quantify the relevance between the $i^{th}$ point's feature $\boldsymbol{f}^i_{pw}$ and the coarse-grained global feature $\boldsymbol{f}_{g\_coarse}$, which can be calculated as:
    \begin{ceqn} \begin{align}
    \label{spf_salient_score}
       a_i=\mathcal{M}\left(\boldsymbol{f}^i_{pw}, {\boldsymbol{f}_{g\_coarse}}\right),
    \end{align} \end{ceqn}
    where $\mathcal{M}$ is the cosine similarity metric function and $i\in[1,n]$. Note that the higher $a_i$ indicates a higher likelihood to be the salient points of the input point cloud $x$. 
    
    Given the salient scores, we select $k_s$ highest salient points, and for each salient point  $p_j$ (where $j\in[1, k_s]$), we further use its $k$-nearest neighbors' ($k$-NNs) features to better describe its local part information as $ \mathcal{N}_{ j} \in \mathbb{R}^{k\times d}$. 
    Next, we concatenate the coarse-grained global feature  $\boldsymbol{f}_{g\_coarse}$ to each local salient part feature  $\mathcal{N}_{ j}$, and perform a $1\times1$ convolution $g(.)$ on the concatenated channels to encode the concatenated features as { $\boldsymbol{f}_{e}^j$}, which is denoted as:
    \begin{ceqn} \begin{align}
    \label{spf_encoded_feature}
    \boldsymbol{f}_{e}^j=g\left([ { \boldsymbol{f}_{g\_coarse}},~\mathcal{N}_{ j}]\right) \in \mathbb{R}^{1\times d},
    \end{align} \end{ceqn} 
    Finally, we obtain the refined global shape descriptor { $\boldsymbol{f}_g$}, by performing a channel-wise Max-pooling function on { $\boldsymbol{F}_{e} = \{\boldsymbol{f}_{e}^j\}$ where $j \in [1, k_s]$}, which can be described as:
    \begin{ceqn} \begin{align}
    \label{spf_output_feature}
       \boldsymbol{f}_{g}=Maxpooling\left(\boldsymbol{F}_{e}\right)  \in \mathbb{R}^{1\times d},
    \end{align} \end{ceqn}

     %================ fig: SCI+ ================%
    \begin{figure*}[t]
    \begin{center}
    \includegraphics[width=1\linewidth]{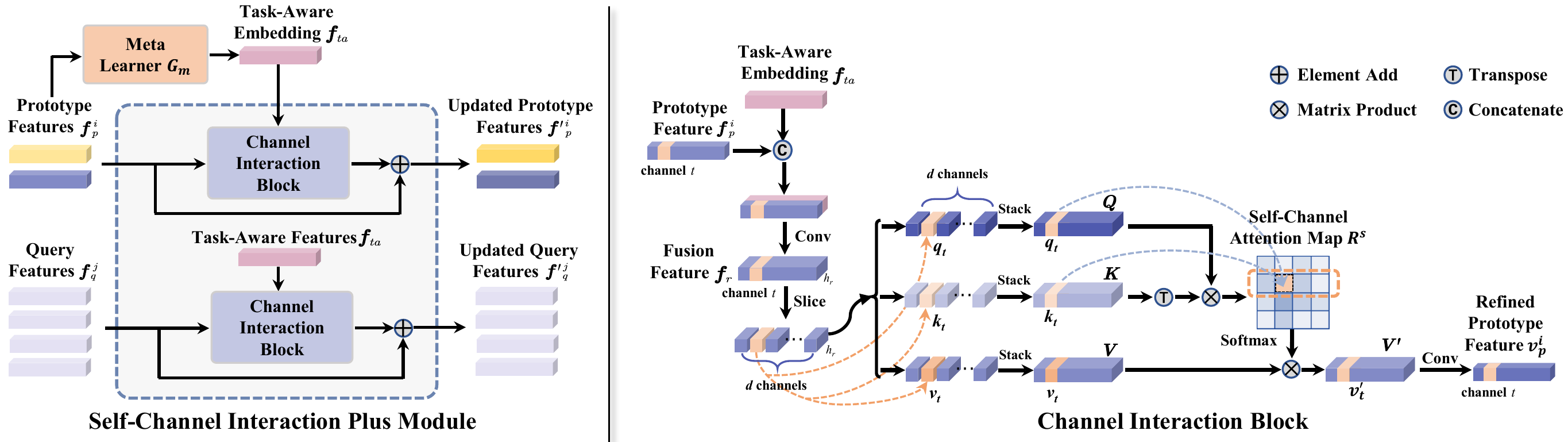}
    \end{center}
    \caption{An illustration of the Self-Channel Interaction Plus (SCI+) module and the Channel Interaction Block (CIB). (1) Firstly, the Meta Learner ${G}_{m}$ takes in the prototype features and outputs the task-aware embedding $\boldsymbol{f}_{ta}$. Then each prototype feature $\boldsymbol{f}_{p}^i$ (or query feature $\boldsymbol{f}_{q}^j$), combined with the task-aware embedding, is fed into the CIB to generate the updated features. (2) In the CIB, we first concatenate the input feature (prototype feature $\boldsymbol{f}_{p}^i$ or query feature $\boldsymbol{f}_{q}^j$) with the task-aware embedding $\boldsymbol{f}_{ta}$ and fuse them with a 1$\times$1 Conv to increase the dimension of each channel. Then we adopt the self-attention mechanism~\cite{self-attention} to obtain a Self-Channel Attention Map $R^s$ and use it to refine the input feature.}
    \label{fig:SCI+_module}
    \vspace{-5pt}
    \end{figure*}
    %================ fig: SCI+ ================%
    
    \subsection{Self-Channel Interaction Plus (SCI+) Module}
    \label{subsec:SCI+_module}
    3D FSL faces the challenge of subtle inter-class differences when the network is trained on a small number of data that may contain fine-grained parts. For example, the samples from the ``chair'' category can come with distinct handles. On the other hand, different channels of a point cloud feature contain various shape-structural information. Our previous work~\cite{ye2022makes} attempted to enhance discriminative parts, by modeling the instance-wise internal structure relationships with the Self-Channel Interaction (SCI) module. 
    
    However, the SCI module only considers the channel attention within one single prototype and ignores the parts shared across other prototypes. For example, ``chair'' and ``stool'' can both have samples with similar handles and cushion structures. In order to increase the network's discriminability for structurally similar objects from different classes, we hope that such inter-class similar structures could be weakened so that the inter-class differences become more obvious and easy to be distinguished.
    
    So, to mine this shared information across different prototypes, we propose the Self Channel Interaction Plus (SCI+) module to elicit structures shared across prototypes within the same task, and then use the attention mechanism to adjust the weight of different channels.
    
    In particular, we first design a Meta Learner $G_m$ to generate the task-aware embedding $\boldsymbol{f}_{ta}$, to model the shared context information among $\boldsymbol{f}^{i}_{p}$. Specifically, the generation of $\boldsymbol{f}_{ta}$ can be described as:
    \begin{ceqn} \begin{align}
    \label{eq:task_aware_embedding}
        \boldsymbol{f}_{ta}= G_m([\boldsymbol{f}^{1}_{p}, \boldsymbol{f}^{2}_{p},...,\boldsymbol{f}^{N}_{p}] ) \in \mathbb{R}^{1\times d},
    \end{align} \end{ceqn}
    where $[\boldsymbol{f}^{1}_{p}, \boldsymbol{f}^{2}_{p},...,\boldsymbol{f}^{N}_{p}]\in \mathbb{R}^{1 \times d \times N}$ is the concatenated feature of the $N$ prototype features, and the $G_m$ is a 1$\times$1 Conv layer performing on the third dimension of the concatenated feature, to encode the $N$ prototype features into a $d$-dimension task-aware embedding $\boldsymbol{f}_{ta}$.
    
    Then, each prototype feature $\boldsymbol{f}^{i}_{p}$ (or query features $\boldsymbol{f}^{j}_{q}$), combined with the task-aware embedding $\boldsymbol{f}_{ta}$, is fed to the Channel Interaction Block (CIB). We adopt the self-attention mechanism~\cite{self-attention} in CIB to obtain a Self-Channel Attention Map $R^s$, and use it to refine the input prototype features (or query features)  {by increasing the weight of discriminative channels and deducing the weight of less discriminative channels containing class-similar information.}
    
    The right part of Fig.~\ref{fig:SCI+_module} illustrates the process of refining prototype feature  $\boldsymbol{f}^{i}_{p}$ in the CIB, where we first concatenate the $\boldsymbol{f}^{i}_{p}$  with the task-aware embedding $\boldsymbol{f}_{ta}$, and then fuse them using a 1$\times$1 Conv layer to encode each channel into $h_r$-dimension to get the fusion feature $\boldsymbol{f}_{r}$:
    \begin{ceqn} \begin{align}
    \label{eq:fusion_feature}
        \boldsymbol{f}_r= Conv([\boldsymbol{f}^{i}_{p}, \boldsymbol{f}_{ta}] ) \in \mathbb{R}^{1 \times d \times h_r},
    \end{align} \end{ceqn}
    where each channel is represented by a $h_r$-dimension vector that integrates the task-related information.  
    
    After that, we slice the $\boldsymbol{f}_{r}$ along the channel dimension $d$. For each channel, we generate a query-vector $\boldsymbol{q}_t\in \mathbb{R}^{1 \times h_r}$, a key-vector $\boldsymbol{k}_t\in \mathbb{R}^{1 \times h_r}$ and a value-vector $\boldsymbol{v}_t\in \mathbb{R}^{1  \times h_r}$ using three liner embedding layers, respectively, to extract salient features in each channel. For efficient computation, we pack these vectors of different channels together into matrices $\boldsymbol{Q} \in \mathbb{R}^{1 \times d \times h_r}$,  $\boldsymbol{K}\in \mathbb{R}^{1 \times d \times h_r}$ and $\boldsymbol{V}\in \mathbb{R}^{1 \times d \times h_r}$, and adopt the matrix-product to obtain the self-channel attention map $R^s$:
    \begin{ceqn} \begin{align}
    \label{eq:self_channel_attention_map}
        R^s = \frac{\boldsymbol{Q}\boldsymbol{K}^T}{\sqrt{h_r}} \in \mathbb{R}^{1 \times d \times d},
    \end{align} \end{ceqn}
    and the weighted value-matrix $\boldsymbol{V'}$ is:
    \begin{ceqn} \begin{align}
    \label{eq:weighted_value_matrix}
        \boldsymbol{V'} = softmax(R^s)\boldsymbol{V}\in \mathbb{R}^{1 \times d \times h_r}.
    \end{align} \end{ceqn}
    
    Finally, we use a 1$\times$1 Conv layer to compress the third dimension of $\boldsymbol{V'}$ from $h_r$ to $1$, and reshape it into a $1\times d$ vector $\boldsymbol{v}_p^i$. After this we can compensate the discarded information by combining the refined features $\boldsymbol{v}_p^i$ with $\boldsymbol{f}^{i}_p$, and get the updated prototype features $\boldsymbol{f'}^i_p$:
    \begin{ceqn} \begin{align}
    \label{eq:res_update}
        \boldsymbol{f'}^i_p = \boldsymbol{v}_p^i + \boldsymbol{f}^i_p  \in \mathbb{R}^{1 \times d},
    \end{align} \end{ceqn}
    where $\boldsymbol{v}_p^i = Conv(\boldsymbol{V'})\in \mathbb{R}^{1 \times d}$. Similarly, the Eq.\ref{eq:fusion_feature}- Eq.\ref{eq:res_update} are also applied to each query feature  $\boldsymbol{f}_q^j$ to get the updated feature $\boldsymbol{f'}_q^j$.

    After being inserted with the SCI+ module, the network can learn more fine-grained discriminative features to separate different classes more obviously, as shown in Fig. \ref{fig:tsne} (c). More visualization analysis of the SCI+ module can be found in Section~\ref{subsec:Visualization_Analysis}.

    \subsection{Cross-Instance Fusion Plus (CIF+) Module}
    \label{subsec:CIF_module}
    Most existing methods~\cite{vinyals2016matching,snell2017prototypical,sung2018learning,garcia2017few} extract support features and query features independently. As a result, there often exists a feature distribution shift (as shown in Fig. \ref{fig:tsne} (a)) between the support set and the query set. Thus, we propose a simple-but-effective Cross-Instance Fusion (CIF) module in our previous conference paper~\cite{ye2022makes}. In this work, we further explore the implicit spatial information and extend the CIF module by considering the global instance-wise correlation.
     
\begin{figure}[tp]
    \begin{center}
    \includegraphics[width=1\linewidth]{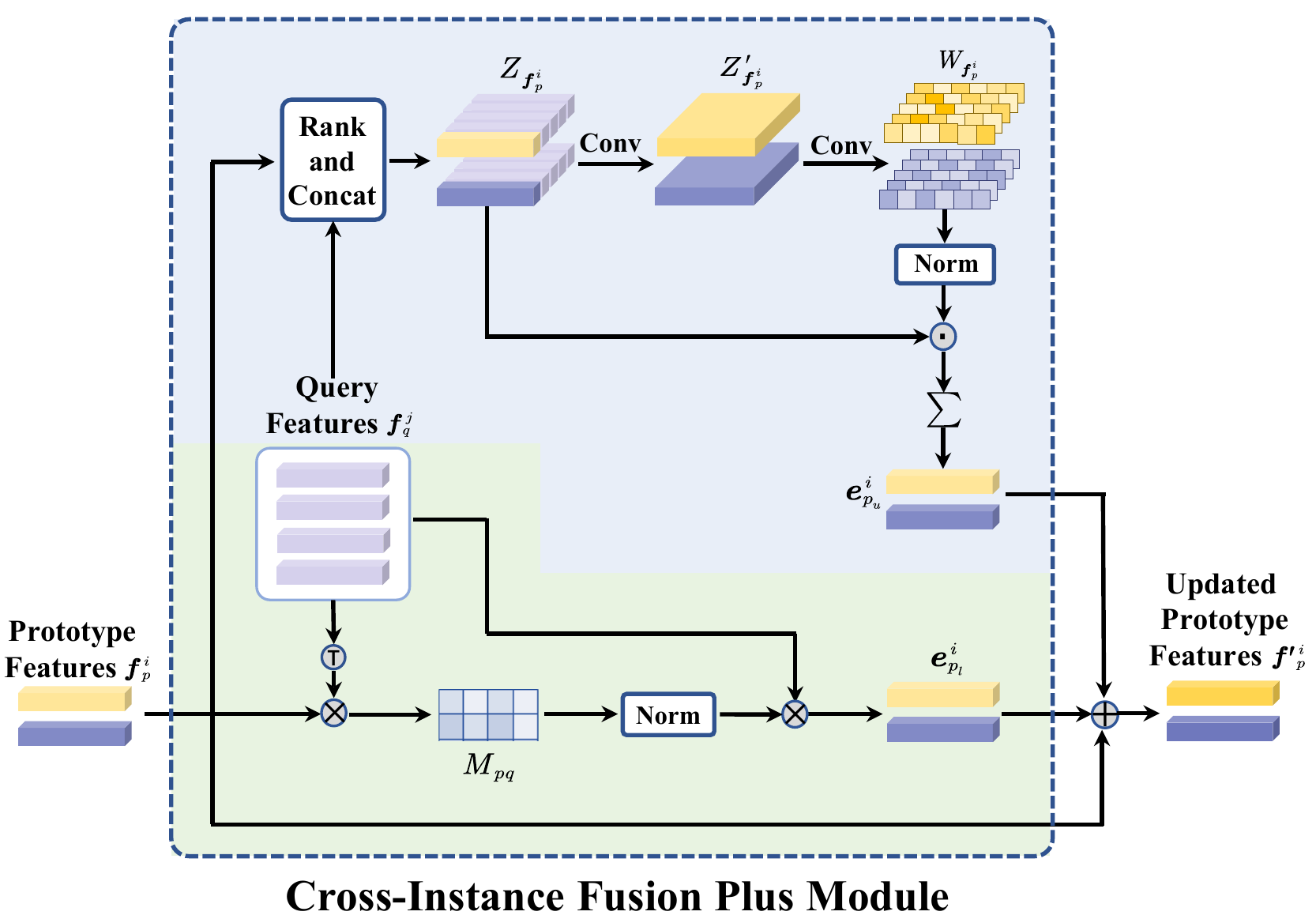} 
    \end{center}
    \caption{An illustration of the Cross-Instance Fusion Plus (CIF+) module. The $\odot$ is the element-wise product, $\oplus$ is the element-wise add, and $\otimes$ is the matrix product. There are two main branches in this module. The upper branch  (masked with blue color) fuses the features based on channel-wise weights, and the lower branch (masked with green color) fuses the features based on the instance-wise weights. Here we only show the process of updating prototype features and the update of query features is similar to this process.} 
    \label{fig:cif_module} \vspace{-10pt}
    \end{figure}

    Specifically, as illustrated in Fig. \ref{fig:cif_module} (here we only show the process of updating prototype features, and the update of query features is similar to the process), there are two main branches in the new CIF module where the upper branch  {(masked with blue color)} fuses the features with channel-wise weights, and the lower branch (masked with green color) fuses the features with the instance-wise weights. Hence, we termed the new CIF module as CIF Plus (CIF+) module.
    
    In the upper branch, we first concatenate each prototype feature $\boldsymbol{f}_p^i$ with its top $K_1$ similar query features based on the cosine distance, and get $Z_{\boldsymbol{f}_p^i}$:
    \begin{ceqn} \begin{align}
    \label{eq:cif_proto_feature_concat}
        Z_{\boldsymbol{f}_p^i}=\left[ \boldsymbol{f}_p^i, \boldsymbol{f}_q^{\left< top1 \right>},..., \boldsymbol{f}_q^{\left< topK_1 \right>} \right]\in \mathbb{R}^{1\times d\times \left( K_1+1 \right)},
    \end{align} \end{ceqn}
    where [$\cdot$] is the concatenation operation, $d$ is the number of feature channels, $K_1 \le N_q$, and $\boldsymbol{f}_q^{\left< top1 \right>}$ represents the query feature having the highest cosine similarity with prototype feature $\boldsymbol{f}_p^i$. 
    
    We then employ two $1\times1$ Conv layers in the concatenation direction to encode the concatenated channels and generate a weight matrix $W_{\boldsymbol{f}_p}$. 
    After that, we update the prototype features by using the weighted sum of $Z_{\boldsymbol{f}_p}$ with  $W_{\boldsymbol{f}_p}$ instead of averaging, which can fuse the channel-wise information flexibly. 
    
    %================ fig: CIF ================%
    %================ fig: CIF ================%
    
    Formally, the weight matrix $W_{\boldsymbol{f}_p^i}$ for $Z_{\boldsymbol{f}_p^i}$ is denoted as:
    \begin{ceqn} \begin{align}
    \label{eq:cif_weight_matrix}
        W_{\boldsymbol{f}_p^i}=f_2\left( f_1\left( Z_{\boldsymbol{f}_p^i} \right) \right)\in \mathbb{R}^{1\times d\times \left( K_1+1 \right)},
    \end{align} \end{ceqn}
    and the fusing feature $\boldsymbol{e}_{p_u}^i$ of the upper branch can be obtained by adding up the $K_1+1$ concatenated features in  $Z_{\boldsymbol{f}_p^i}$ based on  $W_{\boldsymbol{f}_p^i}$:
    \begin{ceqn} \begin{align}
    \label{eq:up_feature}
         \boldsymbol{e}_{p_u}^i=\sum\limits\left(softmax\left(W_{\boldsymbol{f}_p^i} \right) \odot Z_{\boldsymbol{f}_p^i} \right)\in \mathbb{R}^{1\times d}
    \end{align} \end{ceqn} 
    where $f_1\left( \cdot \right)$ is the first $1\times1$ Conv layer encoding the $Z_{\boldsymbol{f}_p^i}$ into a $h$-dim feature interaction ${Z'}_{\boldsymbol{f}_p^i}$, and the second $1\times1$ Conv layer $f_2\left( \cdot \right)$ is designed to adjust the dimension of interaction ${Z'}_{\boldsymbol{f}_p^i}$ to generate the weight matrix $W_{\boldsymbol{f}_p^i}$, and $\odot$ is the element-wise product.

    However, the 1$\times$1 Convolution used in the upper branch only considers the relation within the concatenated channels, yet ignores the correlation between other channels. To further explore the latent spatial information represented by other channels, we extend the CIF module by adding a new branch (masked with green color in Fig.~\ref{fig:cif_module}) that introduces instance-wise weights to finely adjust the features.
    
    Specifically, we first stack the $N$ prototype features and the $N_q$ query features into prototype matrix $\boldsymbol{P}_s\in \mathbb{R}^{N\times d}$ and query matrix respectively $\boldsymbol{Q}_s\in \mathbb{R}^{N_q\times d}$, so that we can easily calculate the cross-instance relation map $M_{pq}$ of prototype and query features with matrix-product:
    \begin{ceqn} \begin{align}
    \label{eq:cross_instance_relation_map}
         M_{pq}= \boldsymbol{P}_s\boldsymbol{Q}_s^T \in \mathbb{R}^{N \times N_q},
    \end{align} \end{ceqn}
    where $N_q= N \times Q $ is the number of total query samples in a meta episode.
    
    Then, we get the fusing features $\boldsymbol{E}_{p_l}$ by combining the query features based on the cross-instance relation map $M_{pq}$:
        \begin{ceqn} \begin{align}
    \label{eq:low_feature}
         \boldsymbol{E}_{p_l}= softmax(M_{pq})\boldsymbol{Q}_s \in \mathbb{R}^{N \times d}.
    \end{align} \end{ceqn} 
    Note that, we implement the Eq.~\ref{eq:low_feature} by matrix-product for efficiency, so the matrix $\boldsymbol{E}_{p_l}$ needs to be reshaped into $N$ $d$-dimension vectors $\boldsymbol{e}_{p_l}^i\in \mathbb{R}^{1 \times d}$, which is the weighted sum of query features based on the $M_{pq}$.
    
    Finally, we update the prototype feature $\boldsymbol{f}_{p}^i$ by adding the fusing features $\boldsymbol{e}_{p_u}^i$ and $\boldsymbol{e}_{p_l}^i$:
    \begin{ceqn} \begin{align}
    \label{eq:cif_update_proto_features}
         \boldsymbol{f'}_{p}^i= \boldsymbol{f}_{p}^i + \boldsymbol{e}_{p_u}^i + \boldsymbol{e}_{p_l}^i \in \mathbb{R}^{1 \times d},
    \end{align} \end{ceqn} 

    The Eq.\ref{eq:cif_weight_matrix}-Eq.\ref{eq:cif_update_proto_features} are similarly applied to the update of query features with prototype features. 
    As shown in Fig. \ref{fig:tsne}(d), the feature distribution gap of different classes is mitigated after using the CIF+ module. More visualization analyses of the CIF+ module are reported in Section~\ref{subsec:Visualization_Analysis}.

%%%%%%%%%   Experiments  %%%%%%%%%
\section{Experiments}
\label{sec:experiment}
\vspace{-5pt}
    This section starts with the details about the newly split benchmark datasets and experimental settings, and then reports the comparison results of the SOTA FSL baseline and our proposed method under different scenarios. After that, extensive ablation studies, insightful analyses, and qualitative visualizations are conducted to evaluate the effectiveness of the proposed modules. 
    \vspace{-10pt}

    \subsection{Implementation Details}
    \label{subsec:setting}
    We take DGCNN~\cite{wang2019dynamic} as the backbone network for feature embedding,  consisting of four EdgeConv layers $(64, 64, 128, 256)$ and an MLP encoded layer. The inputs are a set of meta-tasks containing $N$=5 classes with $K$=1 or $K$=5 support instances and $Q$=15 query instances for each class, termed the “5-way 1-shot 15-query” or “5-way 5-shot 15-query” setting. {We randomly sample 512 points from the CAD model surface or the scanning object for each point cloud example.} We use the Adam optimizer with an initial learning rate of 0.0008 and gamma of 0.5. We adopt the early stop strategy that stops the training when the validation accuracy does not increase in 30 epochs, to reduce overfitting.  The random points jittering and rotating are used to augment data as in ~\cite{qi2017pointnet} during training. Other details of experimental setups are described in the following subsections.

    %================ table: MdelNet+ShapeNet ================%
    \begin{table*}[t] \centering
    \renewcommand{\arraystretch}{1.2}
    \caption{\label{table:synthetic} Comparison results of few-shot point cloud classification (accuracy \%) with 95\% confidence intervals on ModelNet40-FS and ShapeNet70-FS with DGCNN~\cite{wang2019dynamic} as the backbone. We also report the comparison of parameter number (PN), floating-point operations (GFLOPs), and the number of inference tasks per second (TPS) on one NVIDIA 2080Ti GPU. Note that, $^\star$ means the network takes 2D images as input, and $^\dagger$ means taking 2D images with corresponding point cloud instances as input.} 
    \setlength{\tabcolsep}{2.5mm}{
    \begin{footnotesize}
    \scalebox{0.9}{
    \begin{tabular}{c cc cc ccc}
    \hline 
    \multicolumn{1}{c}{
    \multirow{2}{*}{Method}} & 
    \multicolumn{2}{c}{ModelNet40-FS} & 
    \multicolumn{2}{c}{ShapeNet70-FS} &
    \multicolumn{3}{c}{5way-1shot-15query} \\ \cline{2-8} 
    & 5way-1shot   & 5way-5shot 
    & 5way-1shot   & 5way-5shot  
    &  PN      &  GFLOPs    &  TPS \\ \hline 
    Prototypical Net~\cite{snell2017prototypical}      
    & {69.95 $\pm$ 0.67} & {85.51 $\pm$ 0.52} 
    &{69.03 $\pm$ 0.84} & {82.08 $\pm$ 0.72}
    & \tb{0.61M}   & \tb{96.88}   & \tb{12.03}\\
    Relation Net~\cite{sung2018learning}      
    & 68.57 $\pm$ 0.73      & 82.01 $\pm$ 0.53 
    & 67.87 $\pm$ 0.86      & 77.99 $\pm$ 0.70
    & 0.75M   & 97.33   & 11.08\\
    FSLGNN~\cite{garcia2017few}            
    & 61.96 $\pm$ 0.76      & 80.22 $\pm$ 0.55 
    & 66.25 $\pm$ 0.88      & 76.20 $\pm$ 0.77
    & 2.69M  & 101.27  &  10.47\\ 
    Meta-learner~\cite{SachinRavi2017OptimizationAA}
    & 59.08 $\pm$ 0.86      & 76.99 $\pm$ 0.67     
    & 64.53 $\pm$ 0.83      & 74.61 $\pm$ 0.72 
    & 1.42M    & 98.62    & 4.36\\
    MAML~\cite{finn2017model}        
    & 62.57 $\pm$ 0.88      & 77.41 $\pm$ 0.73     
    & 64.39 $\pm$ 0.76      & 74.11 $\pm$ 0.68
    & 1.03M   &  {97.08}   &  9.38\\
    MetaOptNet~\cite{lee2019meta} 
    & 67.05 $\pm$ 0.78      & 85.05 $\pm$ 0.59     
    & 68.27 $\pm$ 0.93      & 81.06 $\pm$ 0.76
    & \tb{0.61M}   & \tb{96.88}   &  5.68 \\ \cline{1-8}
    {S2M2~\cite{PuneetMangla2019s2m2} }
    &  {69.73 $\pm$ 0.64} 	&   {83.25 $\pm$ 0.43} 	
    &  {68.53 $\pm$ 0.73} 	&   {79.71 $\pm$ 0.73}
    &  {0.63M}   &  {97.32} &    {5.23} \\ 
    {Meta-Baseline~\cite{YinboChen2021MetaBaselineES}}
    &   {71.33 $\pm$ 0.34} 	&   {85.27 $\pm$ 0.23} 	
    &   {70.16 $\pm$ 0.41} 	&   {81.08 $\pm$ 0.33} 
    &   {\tb{0.61M}}   &  \underline{97.03}  &    {6.78} \\ 
    {SimpleTrans~\cite{XuLuo2022ChannelIM}} 
    &   \underline{71.44 $\pm$ 0.33} 	 &   {86.78 $\pm$ 0.22} 	 
    &   {69.19 $\pm$ 0.40} 	 &   \underline{83.37 $\pm$ 0.32} 	
    &   {\tb{0.61M}}   &   {98.76}   &    {5.13} \\ \cline{1-8}
    Sharma \etal ~\cite{sharma2020self}   
    &  64.89 $\pm$ 0.82     &  79.59 $\pm$ 0.73 
    &  65.76 $\pm$ 0.72      &  79.19 $\pm$ 0.71
    &  0.73M   &  97.86   &  9.62 \\
    SimpleShot$^\star$~\cite{wang2019simpleshot}
    &  61.87 $\pm$ 0.79      &  78.17 $\pm$ 0.59
    &  61.58 $\pm$ 0.80      &  74.63 $\pm$ 0.63 
    &  1.76M   &  99.18      & \underline{11.55}\\
    LSSB(SimpleShot+SB)$^\dagger$~\cite{stojanov2021using} 
    &  {63.33 $\pm$ 0.75}    &  {76.41 $\pm$ 0.68} 
    &  {64.45 $\pm$ 0.83}    &  {73.77 $\pm$ 0.73}
    &  {-}   &  {-}   &  {-} \\  
    Point-BERT~\cite{yu2021point} 
    &  {69.41 $\pm$ 3.16} & \underline{86.83 $\pm$ 2.03} 
    &  \underline{73.92 $\pm$ 3.60} &  {82.86 $\pm$ 2.92} 
    &   {22.04M}   &   {170.10}   &  {-} \\ 
    {Feng \etal~\cite{HengxinFeng2022EnrichFF}} 
    &   {61.36 $\pm$ 0.41}&	 {73.20 $\pm$ 0.31}	
    &   {65.09 $\pm$ 0.44}	& {75.89 $\pm$ 0.29}
    &   {1.935M}   &    {201.89}   &   {6.03} \\\cline{1-8}
    \multicolumn{1}{c}{\tb{Ours}}                  
    & \textbf{ 81.19 $\pm$ 0.64}  & \textbf{ 89.30 $\pm$ 0.46}  
    & \textbf{ 78.37 $\pm$ 0.79}  & \textbf{ 85.15 $\pm$ 0.70}
    &   \underline{0.63M}   &   {97.37}   &   {7.49}  \\
   \hline 
    \end{tabular}}
    \end{footnotesize}}  \vspace{-5pt}
    \end{table*}
    %================ table: MdelNet+ShapeNet ================%

    \vspace{-5pt}
    \subsection{Experimental Results}
    \label{subsec:Experimental_results}

    \subsubsection{Results on Synthetic Point Clouds}
    \vspace{-5pt}
    We first compare our method  with the aforementioned FSL baselines on two synthetic benchmark datasets, ModelNet40-FS and ShapeNet70-FS. Results reported in the first part of Table~\ref{table:synthetic} prove that our proposed method outperforms other baselines on both datasets by a large margin, by about $10\%$ for 1-shot and $4\%$ for 5-shot. One possible explanation may be that the proposed modules can adjust the feature distribution of the support set and query set by considering channel-level and instance-level correlation, enhancing the distinction between prototypes and query instances in the feature space. Thus the metric-based classifier can better distinguish the boundaries of different classes.
    
    The second part of Table~\ref{table:synthetic} lists the comparison results of recent FSL algorithms. For \tb{S2M2}~\cite{PuneetMangla2019s2m2}, we first pre-train the backbone and classifier head on the base class data with the auxiliary loss (including rotation and exemplar) and cross-entropy loss. After that, we use the support samples to fine-tune the network and predict the query samples' class with a cosine classifier. For Meta-Baseline~\cite{YinboChen2021MetaBaselineES}, we train the network from scratch using the meta-leaning paradigm with a learnable cosine classifier. In SimpleTrans~\cite{XuLuo2022ChannelIM}, we first use the proposed simple transform function to adjust the feature, and then feed them in the Prototypical Net~\cite{snell2017prototypical} to get predictions.

    We also compare our proposed method with four recent 3D learning works that explore low-labeled training data problems in the third part of Table~\ref{table:synthetic}. In particular, for the method proposed by \tb{Sharma \etal}~\cite{sharma2020self}, we first pre-train an embedding network using the self-supervised strategy introduced in their paper, and then fine-tune the whole network and classifier with the labeled support examples. \tb{LSSB}~\cite{stojanov2021using} proposed to use 3D object shape bias to improve the generalization performance of low-shot image classification. To compare with this method, we follow the same experimental settings in ~\cite{stojanov2021using}, which first train the \tb{Simpleshot}~\cite{wang2019simpleshot} on 3D models' multi-view RGB projections with ResNet18~\cite{he2016deep} as the backbone, and then refine the latent embedding space by using 3D object shape bias, denoted as \tb{Simpleshot+SB}~\cite{stojanov2021using}. For \tb{Point-BERT}~\cite{stojanov2021using}, we first pre-train the dVAE model and Transformer encoder on ShapeNet55~\cite{chang2015shapenet}, and then fine-tune the model to the few-shot classification task with the same parameter settings adopted in their paper.
    To compare with the channel-wise attention introduced in the~\cite{HengxinFeng2022EnrichFF}, we directly replace the attention mechanism in the SCI+ module by generating the Q-K-V vector for each point with three FC layers and using them to get the attention score. Then we refine the point-wise feature using this attention score as done in~\cite{HengxinFeng2022EnrichFF}
    
    The comparison results reported in Table~\ref{table:synthetic} show that our proposed network achieves better performance on both benchmark datasets by about $12\%$ for the 1-shot setting and $4\%$ for the 5-shot setting on ModelNet40-FS, and by about $9\%$ for 1-shot setting and $3\%$ for 5-shot setting on ShapeNet70-FS, respectively.
    
    %================ table: ScanObjectNN ================%
    \begin{table*}[t] \centering
    \renewcommand{\arraystretch}{1.2}
    \caption{\label{table:real-world} {Comparison results of few-shot point cloud classification results (accuracy \%) with 95\% confidence intervals on the real-world  point cloud classification dataset, ScanobjectNN-FS, with DGCNN~\cite{wang2019dynamic} as the backbone. $S^i$ denotes the results on split $i$.} }
    \setlength{\tabcolsep}{1.4mm}{
    \begin{footnotesize}
    \scalebox{0.9}{
    \begin{tabular}{c cccc cccc}
   \hline 
    \multicolumn{1}{c}{
    \multirow{3}{*}{Method}}  
    &\multicolumn{8}{c}{ScanObjectNN-FS} 
    \\ \cline{2-9} 
    &\multicolumn{4}{c}{5way-1shot} 
    &\multicolumn{4}{c}{5way-5shot} 
    \\ \cline{2-9} 
    \multicolumn{1}{c}{} 
    & {$S^0$}   & {$S^1$}   & {$S^2$}  &~~Mean~~
    & {$S^0$}   & {$S^1$}   & {$S^2$}  &~~Mean~~
    \\ \hline  
    Prototypical Net~\cite{snell2017prototypical}      
    & 45.10 $\pm$ 0.77  & 44.10 $\pm$ 0.61  & 49.64 $\pm$ 0.61  & 46.28
    & {63.91 $\pm$ 0.50}  & 64.43 $\pm$ 0.50  & 63.49 $\pm$ 0.45  & 63.94 \\
    Relation Net~\cite{sung2018learning}      
    & 45.46 $\pm$ 0.76  & 45.20 $\pm$ 0.58  & 52.26 $\pm$ 0.52  & 47.64
    & 60.45 $\pm$ 0.54  & 52.93 $\pm$ 0.50  & 63.02 $\pm$ 0.47  & 58.80 \\
    FSLGNN~\cite{garcia2017few}            
    & 32.09 $\pm$ 0.78  & 29.37 $\pm$ 0.51  & 25.00 $\pm$ 0.38  & 28.82
    & 38.78 $\pm$ 0.50  & 40.41 $\pm$ 0.50  & 26.77 $\pm$ 0.30  & 35.32 \\ 
    Meta-learner~\cite{SachinRavi2017OptimizationAA}
    & 38.11 $\pm$ 0.47  & 42.05 $\pm$ 0.45  & 41.39 $\pm$ 0.42  & 40.52
    & 51.56 $\pm$ 0.42  & 50.72 $\pm$ 0.40  & 50.96 $\pm$ 0.36  & 51.08 \\
    MAML~\cite{finn2017model}        
    & 39.12 $\pm$ 0.74  & 45.77 $\pm$ 0.62  & 41.02 $\pm$ 0.62  & 41.95
    & 50.22 $\pm$ 0.56  & 51.38 $\pm$ 0.49  & 49.58 $\pm$ 0.28  & 50.39 \\
    MetaOptNet~\cite{lee2019meta}  
    & \underline{45.98 $\pm$ 0.82}  & \underline{48.48 $\pm$ 0.64}	& 58.33 $\pm$ 0.64  & \underline{50.93}	
    & \underline{64.13 $\pm$ 0.51}  & \underline{68.27 $\pm$ 0.49}	& {71.82 $\pm$ 0.47}  & {68.07} \\ \cline{1-9} 
    S2M2~\cite{PuneetMangla2019s2m2}
    &   {43.15 $\pm$ 0.57} 	&   {44.39$\pm$ 0.47} 	&   \tb{64.60 $\pm$ 0.56} 	&   {47.38}
    &   {61.29 $\pm$ 0.64}   &  {65.37 $\pm$ 0.71}   &  {59.52 $\pm$ 0.83} &    {62.06} \\ 
    Meta-Baseline~\cite{YinboChen2021MetaBaselineES}
    &   {45.75 $\pm$ 0.73} 	&   {47.41 $\pm$ 0.59} 	&   {56.77 $\pm$ 0.63} 	&   {49.97} 
    &   {59.35 $\pm$ 0.50} 	&   {67.52 $\pm$ 0.46} 	&   {67.62 $\pm$ 0.43}  &    {64.83} \\ 
    SimpleTrans~\cite{XuLuo2022ChannelIM}
    &   {45.63 $\pm$ 0.69} 	&   {44.22 $\pm$ 0.57} 	&   {54.89 $\pm$ 0.56} 	 &   {48.25} 	
    &   {63.87 $\pm$ 0.49} 	&   {66.07 $\pm$ 0.45} 	&   \tb{74.74 $\pm$ 0.40}   &   \underline{68.23} \\ \cline{1-9}
    Sharma \etal~\cite{sharma2020self}  
    &  {43.75  $\pm$ 0.73}    &  {40.31  $\pm$  0.75}  &  {48.83  $\pm$ 0.59} &  {44.29} 
    &  58.38  $\pm$ 0.63    &  60.31  $\pm$ 0.52  &  {62.42 $\pm$ 0.50} & {60.37}\\
    Point-BERT~\cite{yu2021point}  
    &   {33.22  $\pm$ 2.99}    &  {33.92  $\pm$ 2.21}  &   {38.17  $\pm$ 3.60} &   {35.10} 
    &   {50.32  $\pm$ 2.99}    &   {49.15  $\pm$ 2.53}  &   {53.55 $\pm$ 3.03} &  {51.01}\\
    {Feng \etal~\cite{HengxinFeng2022EnrichFF}} 
    &    {26.43 $\pm$ 0.48}    &   {46.37 $\pm$ 0.77}  &    {49.02 $\pm$ 0.57} &    {40.60} 
    &    {56.10 $\pm$ 0.55}    &    {60.83 $\pm$ 0.47}  &    {52.68 $\pm$ 0.41} &   {56.53}\\\cline{1-9}
    \multicolumn{1}{c}{\tb{Ours}}   
    &\tb{{48.61 $\pm$ 0.75}}	&\tb{{57.96 $\pm$ 0.61}}	&\underline{{62.16 $\pm$ 0.70}}	&\tb{{56.24}}
    &\tb{{65.96 $\pm$ 0.50}}	&\tb{{73.82 $\pm$ 0.45}}	&\underline{{73.11 $\pm$ 0.42}}	&\tb{{70.93}} \\
    \hline 
    \end{tabular}}
    \end{footnotesize}}  \vspace{-10pt}
    \end{table*}
    %================ table: ScanObjectNN ================%
    
    %================ table: larger-way ================%
    \begin{table}[t] \centering 
    \renewcommand{\arraystretch}{1.4}
    \caption{\label{table:larger-way}  Comparisons of the  classification results (accuracy \%) under larger N-way setting on ShapeNet70-FS with DGCNN~\cite{wang2019dynamic} as the backbone. N-w-K-s represent the N-way-K-shot setting.}     
    \setlength{\tabcolsep}{0.7mm}{
    \begin{footnotesize}
    \scalebox{0.8}{
    \begin{tabular}{ccccccc}\hline 
    \multicolumn{1}{c}{\multirow{2}{*}{Method}} & \multicolumn{6}{c}{ShapeNet70-FS} \\ \cline{2-7} 
    \multicolumn{1}{c}{}   & 10w-1s   & 10w-5s & 15w-1s   & 15w-5s  & 20w-1s  & 20w-5s     \\ \hline   
    Prototypical Net~\cite{snell2017prototypical}    	&56.08	&71.29  &49.25	&62.20 &44.03 &58.26\\
    Relation Net~\cite{sung2018learning} 	&53.49	&66.14  &48.68	&58.31 &41.55 &53.78\\ 
    MetaOpt~\cite{lee2019meta}     	&55.68    &68.33       &47.99       &60.72    &43.30 &55.67 \\ 
    Point-BERT~\cite{yu2021point}     	&\underline{ 59.52}    &\underline{ 71.33}       &\underline{ 50.15}       &\tb{65.52}    &\underline{ 48.33} &\tb{61.68} \\ 
    \hline
    \makecell[c]{\tb{Ours}}        	& \tb{{ 62.57}}	&\tb{{ 72.24}}  &\tb{{ 54.04}}	&\underline{{ 64.46}} & \tb{{ 49.68}} & \underline{{ 60.25}} \\ \hline
    \end{tabular}}
    \end{footnotesize}} \vspace{-16pt}
    \end{table}
    %================ table: larger-way ================% 

    \subsubsection{Results on Real-World Point Clouds}
    We further compare the performance of different methods for 3D FSL under a more challenging real-world scenario. Table~\ref{table:real-world} summarizes the results of competing methods on the ScanObjectNN-FS. We can see that the classification performance drops rapidly in the real-world scenario because the data collected from the real world are noisy and cluttered. However, our method also can achieve the top mean accuracy of $56.24\%$ for 1-shot and $70.93\%$ for 5-shot, outperforming most methods on different data splits with different settings. So our proposed method is indicated to be promising and effective for 3D few-shot point cloud classification.
    Note that we can not compare with Simpleshot~\cite{wang2019simpleshot} and LSSB (Simpleshot+SB)~\cite{stojanov2021using} on the ScanObjectNN-FS dataset, because they are image-based methods and the dataset does not provide multi-view 2D RGB projection images.
    
    \vspace{-2pt}
    \subsubsection{Larger N-Way Classification}
    To examine the effect of the proposed network under a practical scenario, we conduct the experiments of the larger N-way setting (where N=10, 15, 20) on ShapeNet70-FS with the backbone network DGCNN~\cite{wang2019dynamic}. As in Table \ref{table:larger-way}, we compare the metric-based methods Prototypical Net~\cite{snell2017prototypical} and Relation Net~\cite{sung2018learning}, the optimization-based method MetaOptNet~\cite{lee2019meta}, and Point-Bert~\cite{yu2021point}.  
     
    The results show that the performances of these methods drop significantly when increasing the number of N. But our approach compares favorably against other methods in most settings. We attribute the results to that larger N-way classification needs to further reduce the intra-class variation and enlarge the inter-class difference, and the proposed SCI+ and CIF+ modules can address these issues to get better performance. Note that, because the Point-BERT~\cite{yu2021point} is pre-trained on the ShapeNet55~\cite{chang2015shapenet}, it has better performance when fine-tuned with more labeled examples (5-shot), but it may be  overfitting when fine-tuning with one labeled example (1-shot). 
    
    %================ table: ablation  ================%
    \begin{table}[t]\centering
    \renewcommand{\arraystretch}{1.3}
    \caption{\label{table:module_ablation} Ablation study of the proposed SPF, SCI+ and CIF+ modules with DGCNN~\cite{wang2019dynamic} as the backbone. We take Prototypical Net~\cite{snell2017prototypical} as the baseline (the first row), ~\cm~ means equipped with that block. } 
    \setlength{\tabcolsep}{1mm}{
    \begin{footnotesize}
    \scalebox{0.9}{
    \begin{tabular}{ccccc cc cc}
    \hline
    \multirow{2}{*}{SPF} 
    &\multirow{2}{*}{SCI+}
    &\multirow{2}{*}{CIF+} 
    &\multicolumn{2}{c}{ModelNet40-FS} 
    &\multicolumn{2}{c}{ShapeNet70-FS}
    &\multicolumn{2}{c}{ScanObjectNN-FS}
    \\ \cline{4-9} 
    &   &   &  5w-1s & 5w-5s & 5w-1s & 5w-5s &5w-1s & 5w-5s \\ \hline 
    &&  &69.95  &85.51  &69.03  &82.08 &46.28  &63.94\\
    \cm &&&  & 71.89  & 87.39  & 69.86  & 83.28 & 47.71  & 67.56 \\
    &\cm &&  & 72.48  & 87.86  & 70.02  & 83.33  & 47.13  & 67.93 \\
    &&\cm&   & 78.54  & 87.89  & 75.65  & 84.44  & 53.11  & 70.20\\ \hline
    \cm& & \cm& & & 73.43  & 88.31  & 70.53  & 83.64  & 48.74  & 68.28 \\
    \cm& & & \cm& &  {79.74}  &   {88.11}  &   {77.93}  &   {84.59}  &   {55.87}  &   {70.53} \\
     & \cm&&  \cm& &   {80.83}  &   {88.61}  &   {77.64}  &   {84.79}  &   {55.38}  &   {70.35} \\
    \cm& & \cm& &\cm& &\tb{81.19}  &\tb{89.30}  &\tb{78.37} &\tb{85.15}	 &\tb{56.24}	 &\tb{70.93}\\      
    \hline
    \end{tabular}}
    \end{footnotesize}} \vspace{-5pt}
    \end{table}
    %================ table: ablation  ================%
    
    %================ table: module_complex  ================%
    \begin{table}[t]
    \centering
    \renewcommand{\arraystretch}{1.1}
    \caption{\label{table:module_complex}  { Complexity analysis of the proposed modules with DGCNN~\cite{wang2019dynamic} as the backbone. We take Prototypical Net~\cite{snell2017prototypical} as baseline (the first row), ~\cm~ means equipped with that block. }}
    \setlength{\tabcolsep}{3mm}{
    \begin{footnotesize}
    \scalebox{0.9}{
    \begin{tabular}{cccccc}
    \hline 
    \multirow{2}{*}{ {SPF}} 
    &\multirow{2}{*}{ {SCI+}}
    &\multirow{2}{*}{ {CIF+}} 
    &\multicolumn{3}{c}{ {5way-1shot-15query}} 
    \\ \cline{4-6} 
    &   &   &   {PN} &  {GFLOPs} &  {TPS}  \\ \hline 
          &&  &   {0.618M}  &  {96.88}  &  {12.03} \\
    \cm &&&  &  {0.618M}  &  {97.00}  &  {8.02} \\
    &\cm &&  &  {0.621M}  &  {97.25}  &  {8.81}  \\
    &&\cm&   &  {0.624M}  &  {97.09}  &  {9.10}  \\
    \cm& & \cm& &\cm& &  {0.627M}  &  {97.37}  &  {7.49} \\    
    \hline 
    \end{tabular}}
    \end{footnotesize}}  \vspace{-9pt}
    \end{table}
    %================ table: module_complex  ================%
    
    %================ table: spf_cia_fsl  ================%
    \begin{table*}[t]  \centering
    \renewcommand{\arraystretch}{1.2}
    \caption{\label{table:spf_cia_fsl}  Comparisons of the classification results (accuracy \%) and improvement, after incorporating the SPF, SCI+ and CIF+ modules into different FSL algorithms with DGCNN~\cite{wang2019dynamic} as the backbone.}    
    \setlength{\tabcolsep}{1mm}{
    \begin{footnotesize}
    \scalebox{0.9}{
    \begin{tabular}{r cc cc cc}
    \hline
    \multicolumn{1}{c}{
    \multirow{2}{*}{Method}} 
    &\multicolumn{2}{c}{ModelNet40-FS} 
    &\multicolumn{2}{c}{ShapeNet70-FS}
    &\multicolumn{2}{c}{ScanObjectNN-FS}
    \\ \cline{2-7} 
    \multicolumn{1}{c}{}   
    & 5w-1s   & 5w-5s  
    & 5w-1s   & 5w-5s 
    & 5w-1s   & 5w-5s  \\ \hline  
    Prototypical Net~\cite{snell2017prototypical}      
    & 69.95  & 85.51  & 69.03  
    & 82.08  & 46.28  & 63.94\\
    Prototypical Net~\cite{snell2017prototypical}+ours 
    &\tb{81.19}  &\tb{89.30}  &\tb{78.37}   
    &\tb{85.15}  &\tb{56.24}  &\tb{70.93} \\
    Improvement
    &+11.24  &+3.79	 &+9.34  
    &+3.07  &+9.96  &+6.99 \\ 
    \hline 
    Relation Net~\cite{sung2018learning} 
    & 68.57  & 82.01  & 67.87  
    & 77.99  & 47.64  & 58.80  \\
    Relation Net~\cite{sung2018learning}+ours 
    & \tb{75.97}  & \tb{85.45}  & \tb{73.91}
    & \tb{83.20}  & \tb{51.40}  &\tb{65.32}  \\
    Improvement
    & +7.40  & +3.44  & +6.04  
    & +5.21  & +3.76  & +6.52 \\ 
    \hline 
    FSLGNN~\cite{garcia2017few} 
    & 61.96  & 80.22  & 66.25  
    & 76.20  & 28.82  & 35.32 \\
    FSLGNN~\cite{garcia2017few}+ours
    &\tb{65.67}  &\tb{83.57}  &\tb{71.20}
    &\tb{80.84}  &\tb{30.32}  &\tb{39.56} \\
    Improvement
    & +3.71	 & +3.35  & +4.95	
    & +4.64	 & +1.50  & +4.24  \\
    \hline 
    Meta-learner~\cite{SachinRavi2017OptimizationAA}  
    & 59.08  & 76.99  & 64.53 
    & 74.61  & 40.52  & 51.08 \\
    Meta-learner~\cite{SachinRavi2017OptimizationAA}+ours 
    & \tb{65.43}   & \tb{79.45}  & \tb{70.14}
    & \tb{80.19}   & \tb{43.89}  & \tb{52.23}  \\
    Improvement
    & +6.35	 & +2.46	 & +5.61	
    & +5.58	 & +3.37	 & +1.15 \\ 
    \hline 
    MAML~\cite{finn2017model}      
    & 62.57  & 77.41  & 64.39 
    & 74.11  & 41.95  & 50.39 \\
    MAML~\cite{finn2017model}+ours  
    & \tb{66.48}  & \tb{79.38}  & \tb{71.24}
    & \tb{77.03}  & \tb{44.34}  & \tb{56.02} \\
    Improvement
    & +3.91	& +1.97   & +6.85	
    & +2.92 & +2.39	  & +5.63 \\ 
    \hline
    MetaOptNet~\cite{lee2019meta}  
    & 67.05   & 85.05  & 68.27
    & 81.06   & 50.93  & 68.07 \\
    MetaOptNet~\cite{lee2019meta}+ours  
    &\tb{76.84}  &\tb{88.13}  &\tb{75.44}
    &\tb{84.07}  &\tb{52.62}  &\tb{69.05} \\
    Improvement
    & +9.79 & +3.08 & +7.17 & +3.01 & +1.69  & +0.98 \\
    \hline
    S2M2~\cite{PuneetMangla2019s2m2}
    &  {69.73} 	&   {83.25}     &   {68.53} 	
    &  {79.71}  &   {47.38}     &	{62.06} \\ 
    S2M2~\cite{PuneetMangla2019s2m2}+ours
    &\tb{75.35}	&\tb{86.23}  &\tb{73.86}
    &\tb{82.37}	&\tb{51.26}	&\tb{68.26} \\
    Improvement
    &  {+5.62}	&  {+2.98}	&  {+5.33}	
    &  {+2.66}	&  {+3.88}	&  {+6.20} \\
    \hline 
    Meta-Baseline~\cite{YinboChen2021MetaBaselineES}
    &   {71.33}   &   {85.27} 	 & {70.16} 	
    &   {81.08}   &   {49.97}	 & {64.83} \\ 
    Meta-Baseline~\cite{YinboChen2021MetaBaselineES}+ours
    &  \tb{81.37}	  &  \tb{88.58} &  \tb{78.88}	  
    &  \tb{84.89}	  &  \tb{6.98}	 &  \tb{71.12} \\ 
    Improvement
    &   {+10.04}  & {+3.31}	 &  {+8.72}	 
    &  {+3.81}	 &  {+7.01}	 &  {+6.29} \\
     \hline 
    SimpleTrans~\cite{XuLuo2022ChannelIM}
    &   {71.44}   &  {86.78}    &   {69.19} 	
    &   {83.37}   &   {48.25}   &	 {68.23} \\ 
    SimpleTrans~\cite{XuLuo2022ChannelIM}+ours
     &  \tb{80.36}	& \tb{87.90}  & \tb{77.92}	
     & \tb{84.20}	& \tb{55.83}	& \tb{70.30}\\
    Improvement
    & {+8.92}	& {+1.12} & {+8.73}	
    & {+0.83}	& {+7.58}	& {+2.07} \\
    \hline
    \end{tabular}}
    \end{footnotesize}}  \vspace{-5pt}
    \end{table*}
    %================ table: spf_cia_fsl  ================%

    \subsubsection{Complexity Comparisons} To study the complexity and cost of the competing FSL methods, we report the parameter number (PN), floating-point operations (GFLOPs), and the average inference tasks per second on one NVIDIA 2080Ti GPU for a 5-way 1-shot task with 15 query examples per class. As shown in the right part of Table~\ref{table:synthetic},
    our proposed network introduces a small computational overhead (about 97.37 GFLOPs) and little parameters increase (about 0.63M) for the 5-way 1-shot task, corresponding to 0.5 GFLOPs and 0.02M relative increase over the Prototypical Net~\cite{snell2017prototypical} baseline, which are less than other baselines.  
    Note that LSSB ( SimpleShot+SB)~\cite{stojanov2021using} takes point cloud and image as input, and it does not provide images on ScanObjectNN-FS, hence we cannot report its complexity and cost on this dataset.  During the testing stage, Point-BERT~\cite{yu2021point} needs to fine-tune the model, and the speed of inference depends on the number of retraining epochs, hence we also cannot report the TPS of Point-BERT precisely. However, the fine-tuning phase in Point-BERT is time-consuming, so its TPS is much lower than other baselines. The attention used in the method~\cite{HengxinFeng2022EnrichFF} needs to generate the Q-K-V vector for each point's feature introducing a larger number of parameters and higher computational cost (about 201.89 GFLOPs).
    
    \subsection{Ablative Analysis}
    \label{subsec:Ablative_Analysis}
    In this section, ablative experiments are first conducted to study the contribution and complexity of each proposed module. Then we analyze the effects of applying the proposed modules to different FSL baselines.
    
    \subsubsection{Ablation Studies} We further perform ablation studies to analyze the contribution of each proposed module. 
    We take Prototypical Net~\cite{snell2017prototypical} with DGCNN~\cite{wang2019dynamic} embedding network as the baseline and incorporate these modules into the baseline to form different combinations. Table~\ref{table:module_ablation} lists the comparison results, in which we can observe that the three proposed components could provide positive impacts and improve the baseline's performance on all datasets. From the perspective of individual modules, the CIF+ module contributes more than SPF and SCI+ modules, suggesting that exploiting cross-instance relationships can generate more information for classification than extracting a single instance feature independently. We also can conclude that the SPF module is more beneficial in the real-world scenario, which obtains about $4\%$ improvement on ScanObjectNN-FS, and $1.8\%$ and $1.2\%$ improvement on ModelNet40-FS and ShapeNet70-FS under the 5-shot setting. By combining these modules, we obtain performance gains on all datasets, by $1.5\%$ - $10\%$. Eventually, the integration of the three proposed modules yields the best performance, increasing the accuracy by about $10\%$ for the 1-shot setting and by $4\%$ for the 5-shot setting, respectively. Moreover, the comparison  results of the SCI+ and CIF+ modules and the SCI and CIF modules can be referred to the Supplementary Material.
    
    \subsubsection{Component Complexity Analysis}
    \label{sec:ab_cost}
    We further analyze the complexity of each proposed module when inserted into the baseline network under the 5-way 1-shot 15-query setting. As shown in Table~\ref{table:module_complex}, the SPF introduces negligible parameters, and the SCI+ and the CIF+ only increase little parameters (about 0.003M and 0.006M) and floating-point operations (about 0.37 GFLOPs and 0.49 GFLOPs). However, they also lead to slight inference speed reduction, about four episodes on average.

    \subsubsection{Applying the proposed modules to Different FSL Baselines}  
    Because the proposed models only impact the point cloud features, they can be easily inserted into the compared FSL baselines mentioned in the Tabel~\ref{table:synthetic}. Therefore, we embed the proposed modules, SPF, SCI+ and CIF+, into metric-based and optimization-based FSL baselines to validate their generalization ability to different networks. Fairly, we take the DGCNN~\cite{wang2019dynamic} as the backbone for feature embedding and train the networks with the same strategy. Table~\ref{table:spf_cia_fsl} reports the improvement of inserting the proposed modules into different baselines. One can observe that there is an approximately $1\sim 11\%$ consistent increase after integrating with the proposed module, which demonstrates that proposed modules can be inserted into both metric-based and optimization-based FSL baselines with significant performance improvement.\vspace{-5pt}

    \subsection{Other Analysis}
    \label{subsec:Insight_Analysis}
    \subsubsection{Impact of Point Cloud Density} 
\begin{figure}[htb]
    	\begin{center}
            \includegraphics[width=0.8\linewidth]{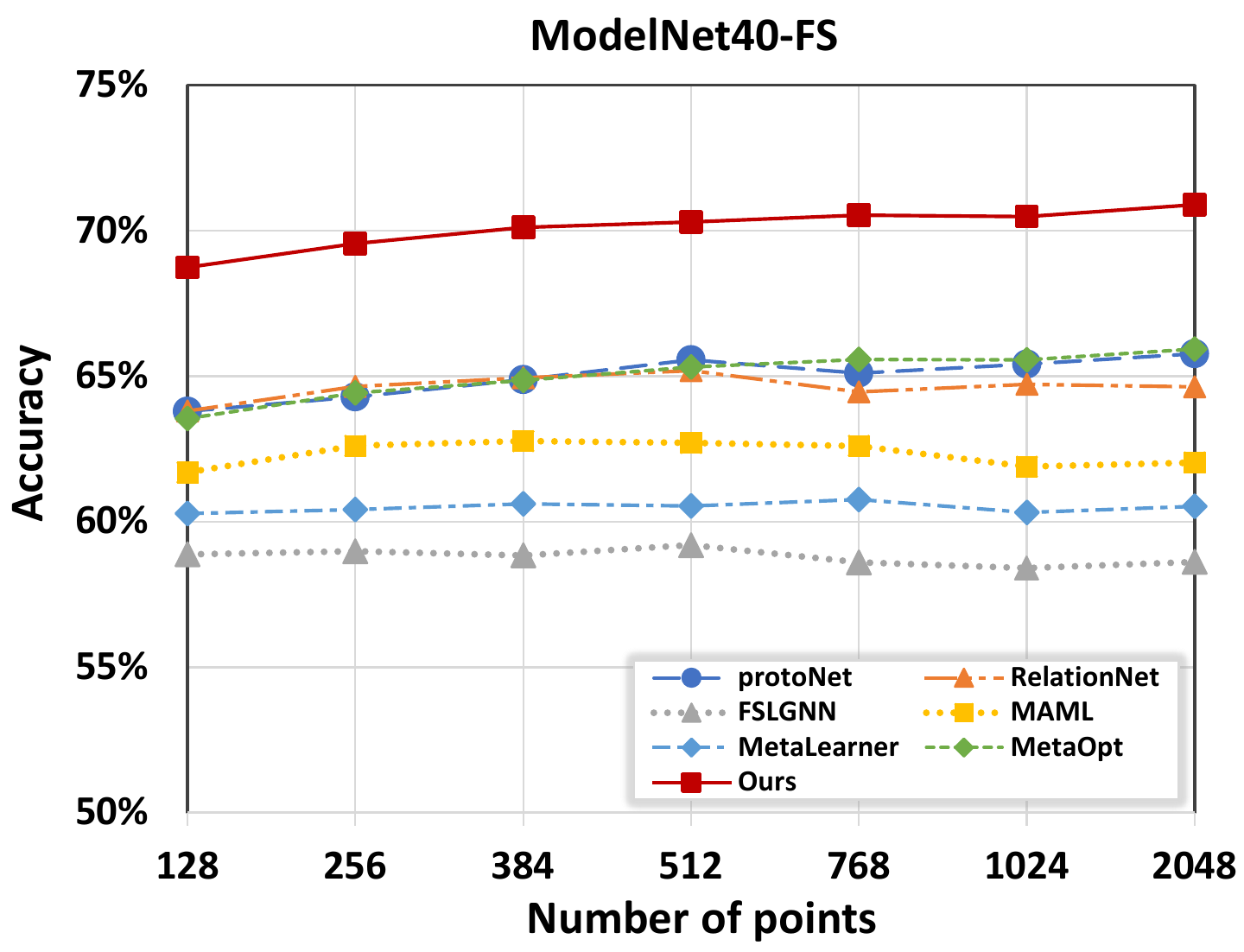} 
    	\end{center}
    	\caption{Comparisons of the classification results with randomly picked points in a point cloud instance on ModelNet40-FS under the 5-way 1-shot setting.}
    	\label{fig:densities}
    % 	\vspace{-5pt}
    \end{figure}

To study the robustness of our proposed network and the FSL baselines with different point cloud densities, we conduct the experiments starting with selecting 128 points and going up to 2048 points for each instance on ModelNet40-FS under the 5-way 1-shot setting. Due to the GPU memory constraint, we take PointNet~\cite{qi2017pointnet} as the backbone. Results in Fig.~\ref{fig:densities} show that the proposed network outperforms other baselines by a large margin under different point cloud density settings. Note that all the baselines have strong robustness for few-shot point cloud classification, even with fewer points. Moreover, for most FSL baselines, a larger number of points only slightly improve the classification results.
    
    %================ fig: n-point  ================%
   %================ fig: n-point  ================%

   %================ fig: cm  ================%
    \begin{figure}[t]
    \begin{center}
    \includegraphics[width=1\linewidth]{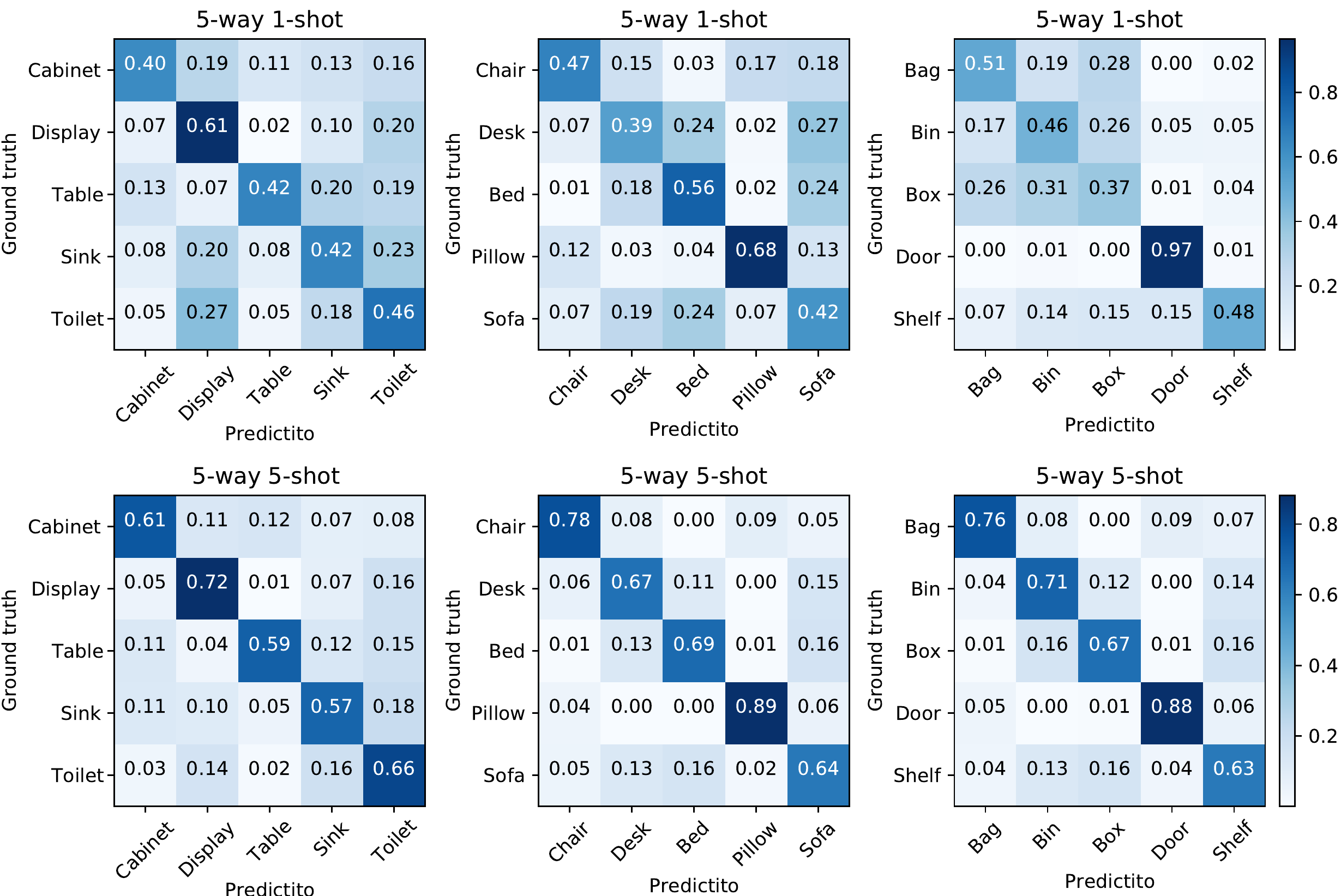} 
    \end{center} 
    \caption{Confusion matrices of classification results only using the coarse-grained global representation on the three splits of ScanObjectNN-FS.}
    \label{fig:spf_cm}
    \end{figure}
    %================ fig: cm  ================%

    %================ table: noise_rate  ================%
    \begin{table}[tb] \centering
    \renewcommand{\arraystretch}{1.3}
    \caption{\label{table:spf_noise_rate}Impact of the SPF module under different noise rates on the ScanObjectNN-FS with DGCNN~\cite{wang2019dynamic} as the backbone.}
    \setlength{\tabcolsep}{1mm}{
    \begin{footnotesize}
    \scalebox{0.9}{
    \begin{tabular}{ccccccc}
    \hline
    \multicolumn{2}{c}{\multirow{2}{*}{ {Setting}}} 
    &\multicolumn{5}{c}{ {Noise Rate}}
    \\ \cline{3-7} 
    \multicolumn{1}{c}{} & \multicolumn{1}{c}{}
    & \multicolumn{1}{c}{ {0.0$\sim$0.2}}
    & \multicolumn{1}{c}{ {0.2$\sim$0.4}}
    & \multicolumn{1}{c}{ {0.4$\sim$0.6}}
    & \multicolumn{1}{c}{ {0.6$\sim$0.8}}
    & \multicolumn{1}{c}{ {0.8$\sim$1.0}}
    \\ \cline{1-7} 
    \multicolumn{1}{c}{\multirow{3}{*}{ {5w1s}}} 
    &\makecell[c]{ {w/o SPF}}
    &  {56.44}  &  {55.41}  &  {51.44}   &  {48.29}  &  \tb{43.61} \\ 
    &\makecell[c]{ {w/ SPF}}
    &  \tb{58.61}  &  \tb{56.83}  &  \tb{52.92}   &  \tb{48.69}  &  {43.32}  \\ 
    &\makecell[c]{ {Improvement}}
    &  {+2.17}  &  {+1.42}  &  {+1.48}   &  {+0.40}  &  {-0.29} \\ \cline{1-7} 
    \multicolumn{1}{c}{\multirow{3}{*}{ {5w5s}}} 
    &\makecell[c]{ {w/o SPF}}
    &  {73.76}  &  {72.89}  &  {67.40}   &  {63.11}  &  {58.05}  \\ 
    &\makecell[c]{ {w/ SPF}}
    &  \tb{76.10}  &  \tb{74.48}  &  \tb{69.23}   &  \tb{64.50}  &  \tb{58.66} \\ 
    &\makecell[c]{ {Improvement}}
    &  {+2.34}  &  {+1.98}  &  {+1.83}   &  {+1.39}  &  {+0.61} \\
    \hline
    \end{tabular}}
    \end{footnotesize}}  
    \end{table}
    %================ table: noise_rate  ================%

    %================ table: classifier  ================%
    \begin{table}[tb] \centering
    \renewcommand{\arraystretch}{1.3}
    \caption{\label{table:classifier}  The comparison results (accuracy \%) of taking different metric functions as the meta classifier.} 
    \setlength{\tabcolsep}{1.5mm}{
    \begin{footnotesize}
    \scalebox{0.9}{
    \begin{tabular}{rcc cccc}
    \hline
    \multicolumn{1}{c}{
    \multirow{2}{*}{Metric}} 
    & \multicolumn{2}{c}{ModelNet40-FS} 
    & \multicolumn{2}{c}{ShapeNet70-FS}
    & \multicolumn{2}{c}{ScanObjectNN-FS}
    \\ \cline{2-7} 
    \multicolumn{1}{c}{}
    & 5w-1s & 5w-5s  & 5w-1s  & 5w-5s & 5w-1s  & 5w-5s\\ \hline  
    \makecell[c]{Cosine\\Similarity}
    &  78.96 & 87.03  & 76.31 & 83.26 & 51.68	 & 66.78\\ \hline
    \makecell[c]{Euclidean\\Distance}
    & \tb{81.19}  &\tb{89.30}  &\tb{78.37} &\tb{85.15}	 &\tb{56.24}	 &\tb{70.93}  \\ 
    \hline
    \end{tabular}}
    \end{footnotesize}}     \vspace{-5pt}
    \end{table}
   %================ table: classifier  ================%
      
    \subsubsection{ Impact of Noise Rate on the SPF Module}
    In the SPF module, salient parts selection relies highly on the coarse-grained global representations. To quantify the quality of the coarse-grained global representation, we first depict the confusion matrices of classification results only using the coarse-grained global representations on the three splits of ScanObjectNN-FS in Fig.~\ref{fig:spf_cm}. One can observe that the coarse-grained global representations also can provide distinguishing information for different classes, especially for the 5-shot setting. 
    
    Then, we further study the performance of the SPF module under different noise rate settings  {to explore the upper limit capability against cluttered background interference.}. Note that we define the noise rate as the ratio of noisy background points to total points in one instance. Results listed in Table~\ref{table:spf_noise_rate} indicate that the SPF module can improve the performance by at least 1.4\% for 1-shot and nearly 2\% for 5-shot when the noise rate is less than 0.6,  {showing good resistance capability of the SPF module to marginal noise and background interference}. However, when the noise rate is higher than 0.6, the SPF module only obtains marginal improvement or even impairs the accuracy for 1-shot. We attribute the results to that a large number of noisy points adversely impact the discrimination of global representations, and noisy backgrounds may also be considered as the potential salient parts, which limits the efficacy of the SPF module.

    \subsubsection{Impact of Global Feature on the SPF Module}
    
    This section investigates the impact of information loss when discarding the non-salient feature points, by removing the global feature in Eq.~\ref{spf_encoded_feature} and simply generating the output feature using only salient parts. Table~\ref{table:spf_global_feature} reports the results on ModelNet40-FS and ScanObjectNN-FS datasets. We can observe that, without the global feature, the performance drops slightly on ModelNet40-FS dataset while more on ScanObjectNN-FS datasets.
    
    %================ table: global_feature  ================%
    \begin{table}[tb] \centering
    \renewcommand{\arraystretch}{1.3}
    \caption{\label{table:spf_global_feature}   { {Ablation results of removing the global features,  where the last row denotes the performance drop on different datasets under different settings after discarding global features.}}} 
    \setlength{\tabcolsep}{2mm}{
    \begin{footnotesize}
    \scalebox{0.9}{
    \begin{tabular}{ccccc}
    \hline
    \multicolumn{1}{c}{\multirow{2}{*}{ { {Method}}}} 
    &\multicolumn{2}{c}{  {ModelNet40-FS}}
    &\multicolumn{2}{c}{  {ScanObjectNN-FS}}
    \\ \cline{2-5} 
    & {5w-1s}&	 {5w-5s}	& {5w-1s}	& {5w-5s}\\ \cline{1-5} 
    \makecell[c]{  {{W / global feature (ours)}}}
    &  \tb{56.24}	& \tb{70.93}	& \tb{81.19}	& \tb{89.30} \\ 
    \makecell[c]{  {{Wo / global feature}}}
    & {55.91}	& {70.91}	& {80.42}	& {88.98}  \\ 
     {performance drop}
    & {-0.33}	& {-0.02}	& {-0.77}	& {-0.32} \\
     \hline
    \end{tabular}}
    \end{footnotesize}}  
    \end{table}
    %================ table: global_feature  ================%
   
    %================ table: Data_Augmentations ================%
    \begin{table}[tb] \centering
    \renewcommand{\arraystretch}{1.3}
    \caption{\label{table:Data_Augmentations}   { {Comparison of different data augmentation methods when applied on the proposed few-shot classification framework on ScanObjectNN-FS.}}} 
    \setlength{\tabcolsep}{3mm}{
    \begin{footnotesize}
    \scalebox{1}{
    \begin{tabular}{ccc}
    \hline
    \multicolumn{1}{c}{\multirow{2}{*}{ { {Method}}}} 
    &\multicolumn{2}{c}{  {ScanObjectNN-FS}}
    \\ \cline{2-3} 
    &  {5w-1s}&	 {5w-5s}	\\ \hline 
     {{No Data-Augment}} &  {54.99}	& {68.08} \\ 
      {{Data-Augment (Ours)}}&   {56.24} & \tb{70.93} \\ 
     {Point-Mixup~\cite{chen2020pointmixup}} & {55.82}	& {70.17} \\
     {Point-Augment~\cite{li2020pointaugment}} & \tb{56.80}	& {70.32} \\
     \hline
    \end{tabular}}
    \end{footnotesize}}  
    \end{table}
    %================ table: Data_Augmentations  ================%
    
    \subsubsection{ {Impact of Different Data Augmentations}}
      {This section studies the impact of different data augmentations when they are used to train the proposed method. Here we compare other two  data augmentations techniques, Point-Mixup~\cite{chen2020pointmixup} and Point-Augment~\cite{li2020pointaugment}. Note that we only augment the data during the training and validating stages using the methods proposed in their paper. Table~\ref{table:Data_Augmentations} lists the few-shot classification results on the ScanObjectNN-FS dataset. One can obverse that the proposed method can work well with different augmentation methods, which have a consistent performance improvement of about 2\% compared with that without augmentation. Especially, the proposed model achieves 56.80\% accuracy in the 1-shot setting if equipped with Point-Augment~\cite{li2020pointaugment}, exceeding the performance of jittering and rotating methods.}

    \subsubsection{Impact of Different Classifiers}
    Here we conduct ablative experiments to study the impact of different classification metric functions for few-shot point cloud classification. Table~\ref{table:classifier} shows the comparison results of using squared Euclidean distance and cosine similarity as classifier $\mathcal{C}_\theta$. We improve the cosine similarity Classifier for a fair comparison by adding a learnable scaling parameter $\tau$. We can observe that the squared Euclidean distance outperforms the cosine similarity by about $2\%$ on all the datasets, so we adopt squared Euclidean distance as the few-shot classification metric for our network. A possible reason is that the squared Euclidean distance is a Bregman divergence, which is suitable for taking the average under point cloud scenario as analyzed by~\cite{snell2017prototypical}.

    %================ fig: distance  ================%
    \begin{figure*}[t]
    \begin{center}
    \includegraphics[width=1\linewidth]{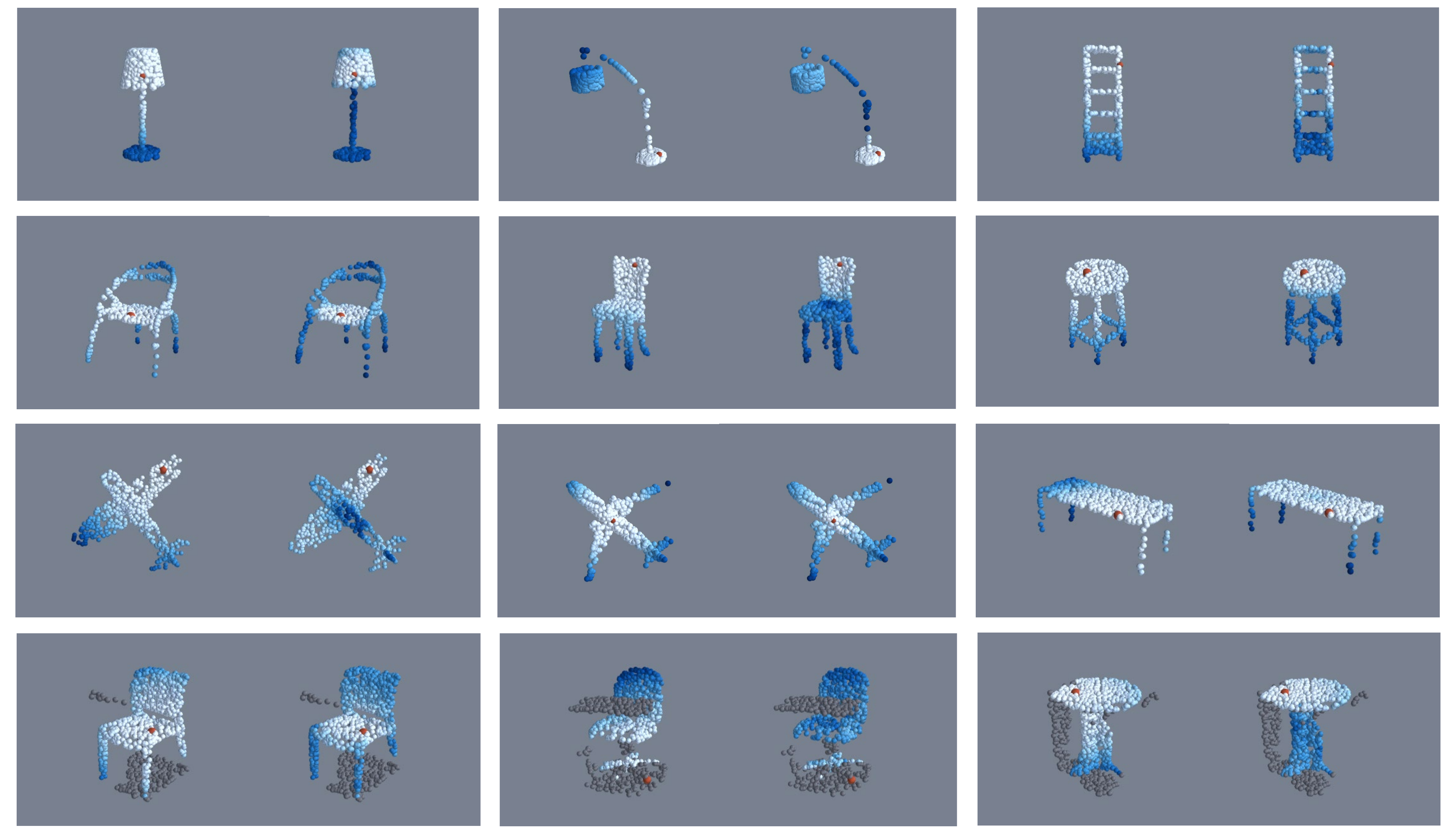} \vspace{-10pt}  
    \end{center}
    \caption{The learned feature space is visualized as a distance from the \tb{RED} point to the rest of the points (\tb{White}: near, \tb{Blue}: far). For clear visualization, we color the attached noisy background as \tb{Gray} points. For each sub-figure, \tb{Left}: Euclidean distance in the input $\mathbbm{R}^3$ space, \tb{Right}: Distance in the learned feature space of our method with DGCNN as the backbone. The semantically similar structures such as the lampshade of a lamp, the shelves of a bookshelf, and the back of a chair are brought close together in the feature space.}
    \label{fig:distance}   
    \end{figure*}
    %================ fig: distance  ================%

    %================ fig: spf_visual  ================%
    \begin{figure*}[t]
    \begin{center}
    \includegraphics[width=0.80\linewidth]{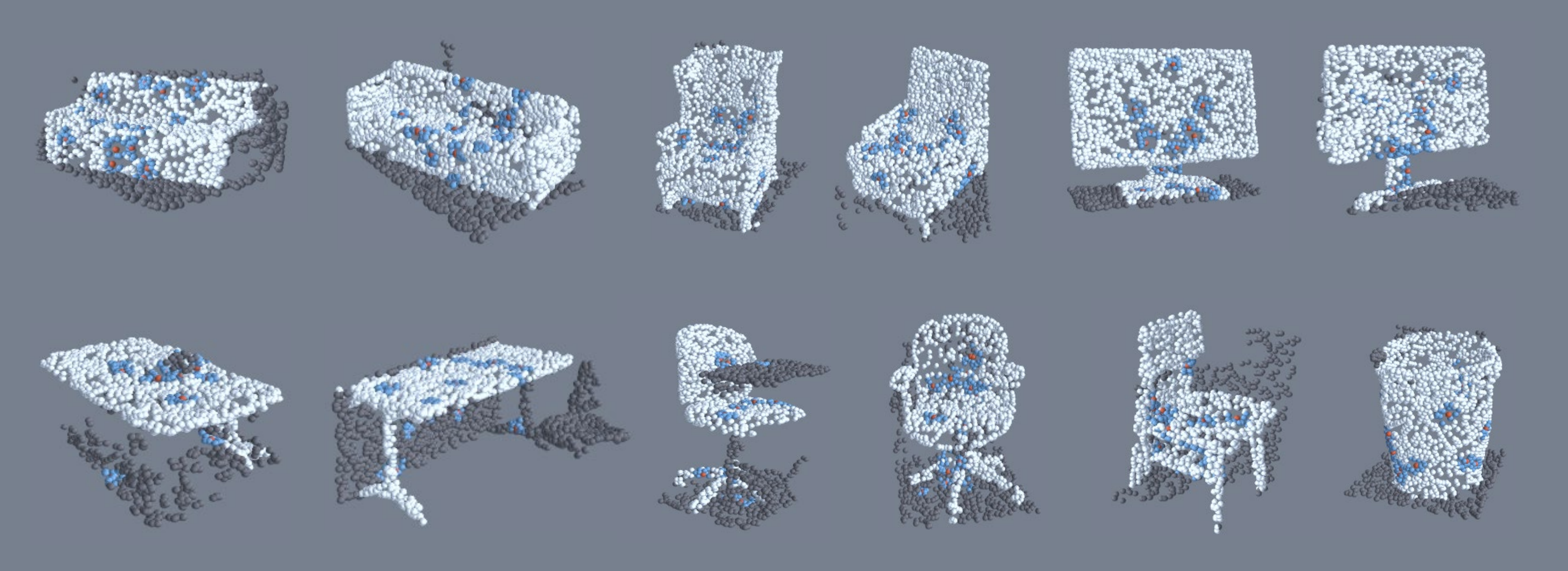} 
    \end{center} 
    \caption{Visualization of Salient Parts in the SPF module. \tb{Red} (better zoom-in to view): the top $k_s$ selected salient points, \tb{Blue}: the k-NN points,  \tb{Gray}: the attached backbone points. }
    \label{fig:visual_spf}  
    \end{figure*}
    %================ fig: spf_visual  ================%

    %================ fig: cia_visual  ================%
    \begin{figure*}[t]
    \begin{center}
    \includegraphics[width=1\linewidth]{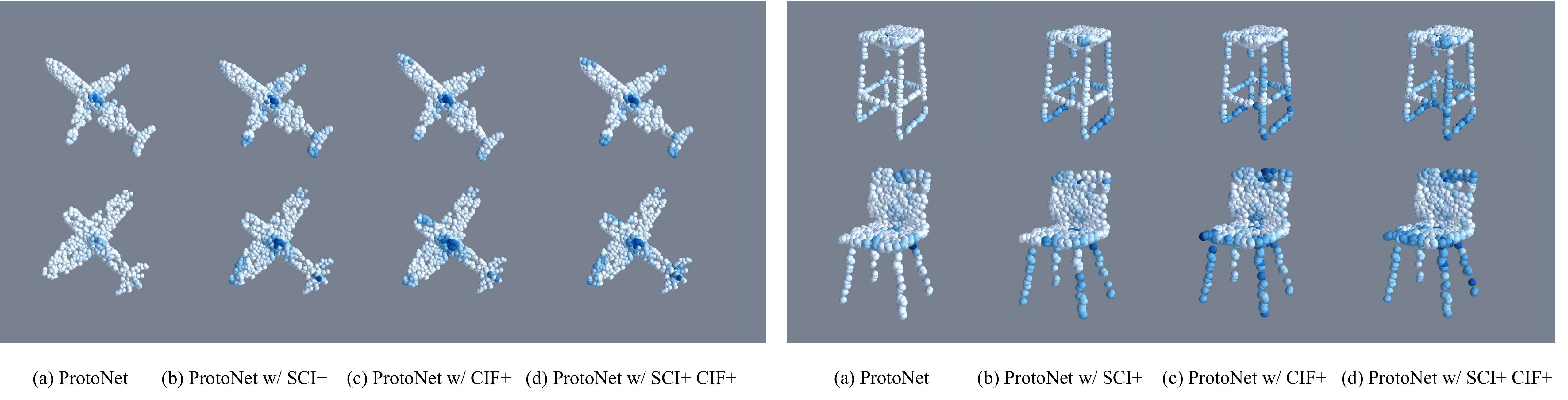} 
    \end{center} 
    \caption{The feature response heatmap of point cloud instances before and after using SCI+ and CIF+ modules. Darker color represents a more intensive feature response in this region. } 
    \label{fig:visual_cia}  
    \end{figure*}
    %================ fig: cia_visual  ================%

    \subsection{Visualization Analysis}
    \label{subsec:Visualization_Analysis}

    \subsubsection{The t-SNE Visualization}
    \label{subsec:t-sne}
    We use t-SNE~\cite{van2008visualizing} to visualize feature distribution under the 5way-5shot-15query setting on ShapeNet70-FS with DGCNN~\cite{wang2019dynamic} as the backbone network. The results are shown in Fig. \ref{fig:tsne}, where (a) corresponds to the features of the Prototypical Net~\cite{snell2017prototypical} baseline, achieving an accuracy of 76.38\%, and (b)-(d) are the results of Prototypical Net~\cite{snell2017prototypical} incorporating the SPF, SCI+, and CIF, achieving 77.53\%, 78.38\%, and 79.30\% respectively, and (e) incorporates both SPF, SCI+ and CIF+ modules, obtaining better performance of 80.03\%. Note that the learning support and query features of Prototypical Net~\cite{snell2017prototypical} are dispersed with huge distribution shifts. After equipping the proposed modules, the features tend to move to the center to better differentiate from other classes, and the distribution shift between support and query sets is mitigated. More t-SNE visualization results can be referred to in Supplementary Material. 

    \subsubsection{Visualization of Feature Space}
    To qualitatively analyze the feature space trained on a few labeled examples, in Fig.~\ref{fig:distance}, we visualize the learned feature space  {output from the backbone network as a distance from the \tb{RED} point to the rest of the points (White: near, Blue: far), as done in work~\cite{sharma2020self}}. The gray points in the last row represent the attached noisy background. For each sub-figure, \tb{Left}: Euclidean distance in the input $\mathbbm{R}^3$ space, \tb{Right}: Distance in the learned feature space of our method with DGCNN as the backbone. One can clearly observe that semantically similar structures such as the lampshade of a lamp, shelves of a bookshelf, and the back of a chair, indicating that the proposed modules can help the backbone to learn a latent feature space in which more discriminative semantic and part information can be captured, although with few labeled training examples.

    \subsubsection{Visualization of Salient Parts in the SPF Module}
    We visualize the salient parts selected by the SPF module in Fig.~\ref{fig:visual_spf}. \tb{Red} points (better zoom-in to view on the computer) represent the salient points and \tb{Blue} points are their k-NNs denoted as the local salient parts. \tb{Gray} points are the noisy background. We show the results of ScanObjectNN-FS instances scanned from the real world. One can observe that  {the proposed model intends to select the points that are more likely to fall on the foreground object and far away from the edges or corners that usually contain background noisy points, indicating that the SPF module could fuse local salient parts more relevant to the shape description to enhance the discrimination of shape embedding while reducing the influences from the noisy background.}
    
    \subsubsection{Visualization of Feature Heatmap of SCI+ and CIF+ Module}
    We further visualize {the feature response heatmap of the SCI+ and the CIF+ module for qualitative evaluation}, by selecting the activated points after the MaxPooling layer and mapping their features into color intensity. {The comparison results shown in Fig.~\ref{fig:visual_cia} depict the change before and after incorporating the SCI+ and CIF+ module.} Note that darker color represents a more intensive feature response in this region.
    We can observe that the SCI+ module pays closer attention to the critical parts of two fine-grained objects, such as the wing and the tail of these two kinds of airplanes, and the leg of different chairs, see Fig.~\ref{fig:visual_cia} (b). While the CIF+ module {focuses on} more diverse regions, like the head of airplanes and the backrest of chairs, see Fig.~\ref{fig:visual_cia} (c). {We can conclude that incorporating the two proposed modules can help to generate more informative features by fusing refined structures and regions}, see Fig.~\ref{fig:visual_cia} (d).
    
%%%%%%%%%   Conclusion  %%%%%%%%%
\section{Conclusion}
\label{sec:conclusion}
    In this work, we {take a closer look at} the 3D FSL problem with extensive experiments and analyses, and design a stronger algorithm for the few-shot point cloud classification task. Concretely, {we first empirically study the performance of recent 2D FSL algorithms when migrating to the 3D domain with different kinds of point cloud backbone networks, and thus construct three comprehensive benchmarks and suggest a strong baseline for few-shot 3D point cloud classification.} Furthermore, to {address} the subtle inter-class differences and high intra-class variance {issuers}, we {come up with a new network, \tb{P}oint-cloud \tb{C}orrelation \tb{I}nter\tb{a}ction (PCIA), with} three plug-and-play modules, namely the Salient-Part Fusion (SPF) Module, the Self-Channel Interaction Plus (SCI+) Module, and the Cross-Instance Fusion Plus (CIF+) Module. {These modules can be easily inserted} into different FSL algorithms with significant performance improvement for 3D FSL. We validate the proposed network on three 3D FSL benchmark datasets, including  ModelNet40-FS,  ShapeNet70-FS, and ScanObjectNN-FS, on which our network achieves state-of-the-art performance.

%%%%%%%%%   Acknowledgements  %%%%%%%%%
\begin{acknowledgements}
  This work is supported by the National Natural Science Foundation of China (No. 62071127, U1909207 and 62101137), Zhejiang Lab Project  (No.2021KH0AB05), the Agency for Science, Technology and Research (A*STAR) under its AME Programmatic Funding Scheme (Project A18A2b0046), A*STAR Robot HTPO Seed Fund (Project C211518008), and Economic Development Board (EDB) OSTIN STDP Grant (Project, S22-19016-STDP).
\end{acknowledgements}

% \clearpage
\bibliographystyle{spbasic}      % basic style, author-year citations
\bibliography{3dfsl_main}   % name your BibTeX database

\clearpage
\tb{\LARGE{\\Supplementary Material}}

\renewcommand\thesection{\Alph{section}}
\setcounter{section}{0}

\section{Details about 3D Backbones}
 \vspace{-5pt}
    To study the influence of different backbone architectures on FSL, we select three types of current state-of-the-art 3D networks as the support backbones for feature extraction. In this section, we will introduce more adopting details about the backbones employed in Section 3. Note that the input point cloud instances consist of 512 points with 3d coordinates and the backbones output a feature vector with 1024 dimensions.

    \tb{Pointwise MLP Networks:}  
    
    \textbf{PointNet}~\cite{qi2017pointnet} contains five MLP layers (64,64,64, 128,1024) with learnable parameters, and batch normalization is used for all MLP layers with ReLU. After that, we use the maxpooling function to aggregate a global feature vector. Note that we remove the transform layers in the original PointNet~\cite{qi2017pointnet} framework for simplicity and efficiency. \textbf{PointNet++}~\cite{qi2017pointnet++} consists of 3-level PointNet Set Abstractions with single scale grouping (SSG), which have the same setting in ~\cite{qi2017pointnet++}. We remove the fully connected (FC) layers and take the last PointNet Set Abstraction's output as the global feature vector. 

    \tb{Convolution Networks:} 
    
    There are 4 X-conv layers (48,96,192,384) with ReLU in \textbf{PointCNN}~\cite{li2018pointcnn}. The last X-conv layer outputs a 384-dimension feature vector and we use 2 FC layers (512,1024) to extend the dimension to 1024 for fair comparisons. \textbf{RSCNN}~\cite{liu2019relation} contains 3 RC-Conv layers (128,512, 1024) with single scale neighborhood (SSN) grouping. Other settings of RC-Conv are the same with~\cite{liu2019relation}. \textbf{DensePoint}~\cite{liu2019densepoint} contains 3 P-Conv layers, 2 P-Pooling layers, and 1 global pooling layer. The settings of these layers are the same with~\cite{liu2019densepoint}. We remove the FC layers in RSCNN and DensePoint for outputting the 1024-dimension global feature vectors too.

    \tb{Graph-based Network:}  
    
    \textbf{DGCNN}~\cite{wang2019dynamic} is the embedding network of our baseline for 3DFSL, consisting of 4 EdgeConv layers (64,64,128,256). The outputs of each EdgeConv will be concatenated as a 512-dimension feature map. Then the feature map will be fed into an MLP layer to extend its dimension to 1024. At last, a max-pooling function is used to aggregate the global features and outputs a 1024-dimension feature vector.

\section{Comparisons with the Conference and Journal version}
    This section summarizes the differences between the conference and journal versions, including the network designs, performance, and complexity.

    \subsection{Comparisons of Network Designs}
\begin{figure*}[thb]
    	\begin{center}
            \includegraphics[width=1\linewidth]{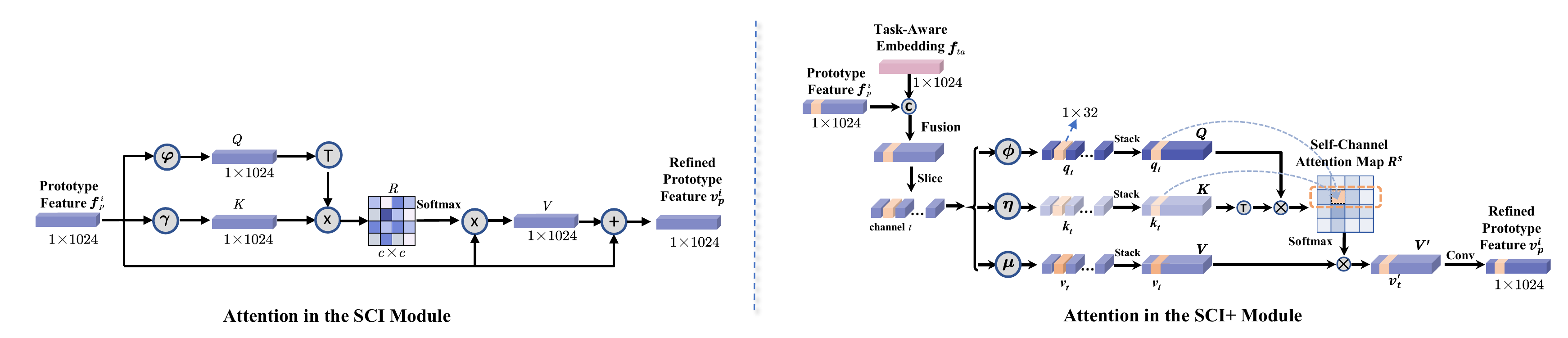} 
    	\end{center} 
    	\caption{Two Different attention mechanisms of the SCI and the SCI+ module.}
    	\label{fig:sci_sci+} 
    	
    \end{figure*}

    Here we simply summarize the network's differences between SCI v.s. SCI+ and CIF v.s. CIF+. Concretely, there are two main differences between the SCI module and the SCI+ module. The first one is that the SCI+ module introduces a meta-learner (that is not included in the SCI module) to generate task-aware embeddings, which can offer task-relevant information when using the attention mechanism to adjust the weight of different channels. The second one is the way to generate the attention score. As illustrated in Fig.~\ref{fig:sci_sci+}, the SCI module learns the Q-K-V vector from the whole global feature which is high dimensional (i.e.1024-dim in the SCI module), causing the parameter numbers of the FC layers to be considerably large. However, the SCI+ module uses task-aware embeddings to extend the dimension of each channel (32-dim in the SCI+ module) and then learns the Q-K-V vector for each channel with smaller-sized FC layers, which can significantly reduce the number of parameters. 

\begin{figure*}[thb]
    	\begin{center}
            \includegraphics[width=1\linewidth]{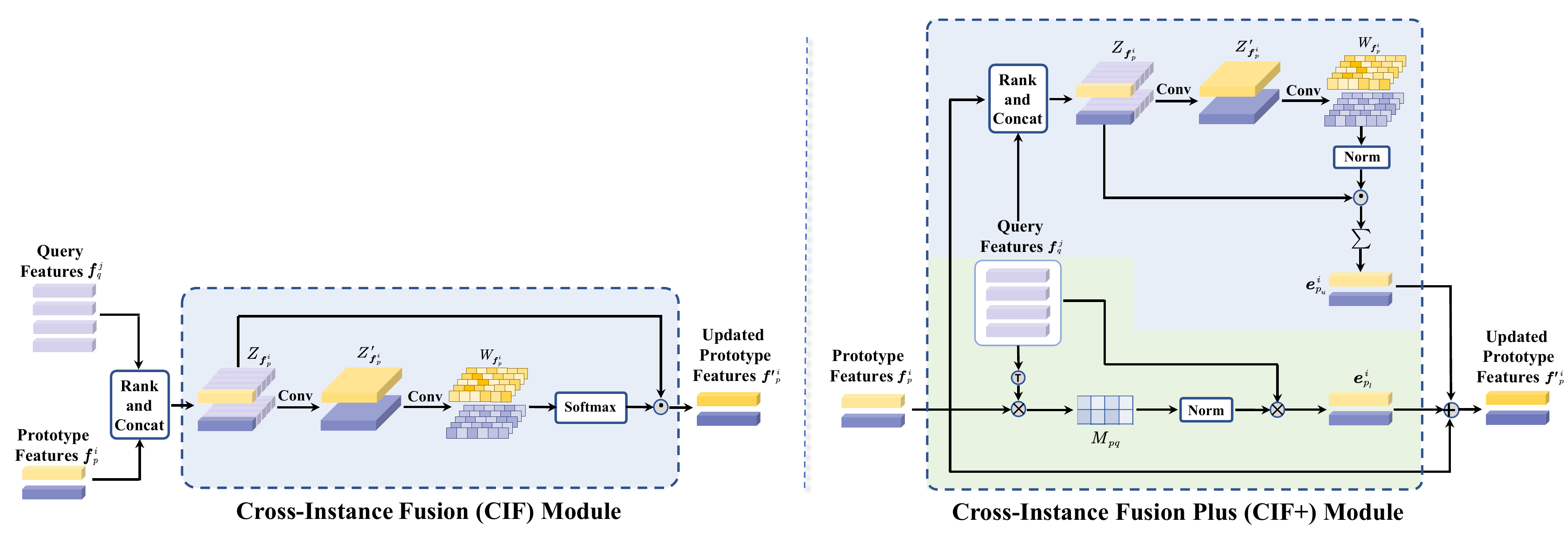}
    	\end{center} 
    	\caption{ The different network designs of the CIF and the CIF+ module}
    	\label{fig:cif_cif+} 
    	
    \end{figure*}

    For the CIF+ module, we mainly extend the CIF module with a new branch that considers the global instance-wise correlation. In this branch, we get a cross-instance relation map to refine the weight of different channels further, helping to better fuse the prototype features and query features, as illustrated in Fig.~\ref{fig:cif_cif+}. 

    Moreover, in Table~\ref{table:module_comparison}, we compare the results of the two new proposed modules (SCI+ and CIF+) with their original design (SCI and CIF) in~\cite{ye2022makes}. ${\ast}$ is the result of our conference version~\cite{ye2022makes}, and ${\dagger}$ is the result of our new proposed method in this paper. We can observe that both newly designed modules (SCI+ and CIF+) can improve the few-shot classification on all settings by about 3\% for 1-shot and 1\% for 5-shot. Furthermore, the new method proposed in this paper considerably exceeds the method of our conference version by about 5\% for 1-shot and 2\% for 5-shot, respectively. 

    %================ fig: tsne2  ================%
    %================ fig: tsne2  ================%
    
    %================ fig: tsne2  ================%
    %================ fig: tsne2  ================%
    
    %================ table: ablation  ================%
    \begin{table}[t]\centering
    \renewcommand{\arraystretch}{1.1}
    \caption{\label{table:module_comparison}The comparison results of the two newly designed modules (SCI+ and CIF+) and their original version (SCI and CIF). ${\ast}$ is the result of our conference version~\cite{ye2022makes}, and ${\dagger}$ is the result of our new proposed method in this paper. } 
    \setlength{\tabcolsep}{1mm}{
    \begin{footnotesize}
    \scalebox{0.95}{
    \begin{tabular}{c cccc}
    \hline
    \multicolumn{1}{c}{\multirow{2}{*}{Method}}
    &\multicolumn{2}{c}{ModelNet40-FS} 
    &\multicolumn{2}{c}{ShapeNet70-FS}
    \\ \cline{2-5} 
    &5way-1shot & 5way-5shot &5way-1shot & 5way-5shot \\ \hline 
    SCI  &70.51  &85.96  &69.37  &82.32 \\
    \tb{SCI+} &\tb{72.48}  &\tb{87.86}  &\tb{70.02}  &\tb{83.33}  \\\hline
    
    CIF  &74.67  &86.69  &72.32  &82.73 \\
    \tb{CIF+} &\tb{78.54}  &\tb{87.89}  &\tb{75.65}  &\tb{84.44} \\\hline
    Ours$^{\ast}$ &75.70  &87.15  &73.57  &83.24 \\
    \tb{Ours$^{\dagger}$} &\tb{81.19}  &\tb{89.30}  &\tb{78.37}  &\tb{85.15} \\
    \hline
    \end{tabular}}
    \end{footnotesize}} 
    \end{table}
    %================ table: ablation  ================%

\subsection{Comparisons of Performance}

    This section compares the  few-shot classification results between the journal and conference version~\cite{ye2022makes} in Table~\ref{table:c_J}. One can observe that the newly designed modules have improved all the methods listed in Table~\ref{table:c_J} compared to the conference version.

    %================ table: spf_cia_fsl  ================%
    \begin{table}[t]  \centering
    \renewcommand{\arraystretch}{1.2}
    \caption{\label{table:c_J}Comparisons of the classification results (accuracy \%) and improvement, after incorporating the SPF, SCI+ and CIF+ modules into different FSL algorithms with DGCNN~\cite{wang2019dynamic} as the backbone.  ${\ast}$ is the result of our conference version~\cite{ye2022makes}, and ${\dagger}$ is the result of our new proposed method in this paper.}
    \setlength{\tabcolsep}{3mm}{
    \begin{footnotesize}
    \scalebox{1}{
    \begin{tabular}{r cc cc}
    \hline
    \multicolumn{1}{c}{
    \multirow{2}{*}{ {Method}}} 
    &\multicolumn{2}{c}{{ModelNet40-FS}} 
    &\multicolumn{2}{c}{{ShapeNet70-FS}}
    \\ \cline{2-5} 
    \multicolumn{1}{c}{}   
    &  {5w-1s}   &  {5w-5s}  
    &  {5w-1s}   &  {5w-5s} \\ \hline  
     {Prototypical Net{$^{\ast}$}}    
    &  {75.70} & {87.15} & {73.57} & {83.24} \\
     {Prototypical Net{${^\dagger}$}} 
    & {\tb{81.19}}  & {\tb{89.30}}  & {\tb{78.37}} & {\tb{85.15}}  \\
     {mprovement}
    & {+5.49}	& {+2.15}	& {+4.8}	& {+1.91}\\  \hline 
     {Relation Net$^{\ast}$}  
    & {70.55} & {83.59} & {68.67} & {78.60}  \\
     {Relation${^\dagger}$}  
    &  {\tb{75.97}}  &  {\tb{85.45}}  &  {\tb{73.91}} &  {\tb{83.20}}  \\
     {Improvement}
    & {+5.42}	& {+1.86}	& {+5.24}	& {+4.6}\\ \hline 
     {FSLGNN$^{\ast}$} 
    & {63.81} & {83.57} & {67.40} & {78.62} \\
     {FSLGNN${^\dagger}$}
    & {\tb{65.67}}  & {\tb{84.67}}  & {\tb{71.20}} & {\tb{80.84}}  \\
     {Improvement}
    & {+1.62}	& {+1.1}	& {+2.74}	& {+1.59} \\ \hline 
     {Meta-learner$^{\ast}$} 
    &  {60.55} & {77.30} & {65.61} & {75.01} \\
     {Meta-learner${^\dagger}$}
    &  {\tb{65.43}}   &  {\tb{79.45}}  &  {\tb{70.14}} &  {\tb{80.19}}  \\
     {Improvement}
    & {+4.88}	& {+2.15}	& {+4.53}	& {+5.18} \\  \hline 
     {MAML$^{\ast}$}    
    &  {63.32} & {78.29} & {65.25} & {75.03} \\
     {MAML${^\dagger}$}  
    &  {\tb{66.48}}  &  {\tb{79.38}}  &  {\tb{71.24}} &  {\tb{77.03}} \\
     {Improvement}
    & {+3.16}	& {+1.09}	& {+5.99}	& {+2.00} \\   \hline
    { MetaOptNet$^{\ast}$}   
    &  {74.70}	& {87.10}	& {72.82}	& {83.08} \\
     {MetaOptNet${^\dagger}$}  
    & {\tb{76.84}}  & {\tb{88.13}}  & {\tb{75.44}} & {\tb{84.07}}  \\
     {Improvement}
    & {+2.14}	& {+1.03}	& {+2.62}	& {+0.99} \\  \hline
    \end{tabular}}
    \end{footnotesize}}  \vspace{-5pt}
    \end{table}
    %================ table: spf_cia_fsl  ================%

\subsection{Comparisons of Complexity}

    {This section compares the computational cost between the conference (SCI, CIF) and journal (SCI+, CIF) versions. The experiment is conducted under the 5-way 1-shot 15-query setting with DGCNN~\cite{wang2019dynamic} as the backbone network. The results listed in Table~\ref{table:complex} show that the parameter number (PN) of the SCI+ module is significantly smaller than that of the SCI module (about 1.859 M) due to the improved attention mechanism. The CIF+ only increases a minor number of parameters (about 0.014 M) and floating-point operations (about 0.006 GFLOPs). Therefore, the total PN of the journal version method decreases considerably while only introducing little floating-point operations.} 

   %================ table: Complexity  ================%
    \begin{table}[t]
    \centering
    \renewcommand{\arraystretch}{1.1}
    \caption{\label{table:complex}  {  {Comparison of the computational cost betweent the conference ${\ast}$ and journal version ${\dagger}$ with DGCNN~\cite{wang2019dynamic} as the backbone network. PN: parameter number. GFLOPs: the number of floating-point operations.  }}}
    \setlength{\tabcolsep}{5mm}{
    \begin{footnotesize}
    \scalebox{1}{
    \begin{tabular}{ccc}
    \hline 
     {Method} &  {PN} &   {GFLOPs}   \\ \hline 
     {SCI / SCI+}	 & {2.48M / 0.621M} & {97.16 / 97.25}\\
     {CIF / CIF+}	& {0.61M / 0.624M}	 & {97.03 / 97.09}\\
     {Ours$^{\ast}$ / Ours$^{\dagger}$}	& {2.54M / 0.627M}  & {97.19 / 97.37}\\
    \hline 
    \end{tabular}}
    \end{footnotesize}}  \vspace{-9pt}
    \end{table}
    %================ table: Complexity  ================%

\section{Extra Experimental Results}

    %================ table: Pointnet+fine-tune ================% 
    \begin{table*}[thb] \centering
    \renewcommand{\arraystretch}{1.3}
    \caption{\label{table:fine_tune} {The comparison results of fine-tuning baseline on ModelNet40-FS, ShapeNet70-FS and the first split of ScanObjectNN-FS, with PointNet~\cite{qi2017pointnet} as the backbone. We report the results of two fine-tuning strategies for different fine-tuning epochs. We test the network on 700 meta-testing episodes randomly sampled from the testing set and report the results with $95\%$ confidence intervals.}}
    \setlength{\tabcolsep}{2mm}{
    \begin{footnotesize}
    \scalebox{0.85}{
    \begin{tabular}{c c cc cc cc}
    \hline
    \multicolumn{1}{c }{\multirow{2}{*}
    {\makecell[c]{Fine-Tuning \\ Strategies}}} & 
    \multicolumn{1}{c }{\multirow{2}{*}
    {\makecell[c]{Fine-Tuning \\ Epochs}}} & 
    \multicolumn{2}{c }{ModelNet40-FS} & 
    \multicolumn{2}{c }{ShapeNet70-FS} &
    \multicolumn{2}{c }{ScanObjectNN-FS ($S^1$)} \\ \cline{3-8} 
    & & 5way-1shot & 5way-5shot & 5way-1shot 
    & 5way-5shot & 5way-1shot   & 5way-5shot   \\ \hline
    \multirow{4}{*}{FC Layers}
    &20 epochs &43.02$\pm$0.90 &50.47$\pm$0.94 &37.61$\pm$0.92 &51.15$\pm$1.03 &30.71$\pm$0.56 &35.96$\pm$0.58       \\
    &40 epochs 
    &60.03 $\pm$0.84  &78.25$\pm$0.63 &65.24$\pm$0.86 &78.91$\pm$0.73 &40.00$\pm$0.70 &56.83 $\pm$0.49       \\
    &60 epochs 
    &58.86$\pm$0.88 &80.64$\pm$0.56 &66.20$\pm$0.86 &\underline{80.18$\pm$0.74} &40.66$\pm$0.72 &58.45$\pm$0.49       \\
    &80 epochs 
    &\underline{63.18$\pm$0.83} &\underline{81.32$\pm$0.59} &65.39$\pm$0.89 &80.02$\pm$0.72 &40.86$\pm$0.67 &\underline{58.72$\pm$0.51}       \\ \hline
    \multirow{4}{*}{\makecell[c]{Backbone \\ + \\ FC Layers}}
    &20 epochs &36.54$\pm$0.83 &44.68$\pm$0.86 &39.07$\pm$0.93 &55.93$\pm$1.00 &30.92$\pm$0.57 &37.05+ 0.56       \\
    &40 epochs &56.42$\pm$0.88 &76.40$\pm$0.65 &65.16$\pm$0.98 &78.48$\pm$0.77 &40.72$\pm$0.68 &55.40$\pm$0.51       \\
    &60 epochs &59.13$\pm$0.83 &79.08$\pm$0.62 &66.13$\pm$0.90 &79.99$\pm$0.72 &\underline{41.21$\pm$0.72} &57.53$\pm$0.51       \\
    &80 epochs &59.81$\pm$0.79 &79.41$\pm$0.57 &\underline{66.53$\pm$0.90} &79.77$\pm$0.73 &40.90$\pm$0.67 &57.07$\pm$ 0.50       \\ \hline
    \multicolumn{2}{c}{Ours}
    &\tb{69.68$\pm$0.40} &\tb{83.19$\pm$0.56} &\tb{70.89$\pm$0.47} &\tb{82.68$\pm$0.63} &\tb{43.56$\pm$0.75} &\tb{62.30$\pm$0.60} \\ \hline
    \end{tabular}}
    \end{footnotesize}}  
    \end{table*}
    %================ table: Pointnet+fine-tune ================%
    
  \subsection{Comparisons with the Fine-Tuning Baseline}
   Here we further study the performance of the fine-tuning baseline.
   Firstly, we pre-train a classification network on the base classes data, which takes PointNet~\cite{qi2017pointnet} as the backbone and two FC layers as the classifier.  Then we fine-tune the pre-trained network on the labeled support examples and evaluated on the query examples of each meta-testing episode. Table~\ref{table:fine_tune} reports two fine-tune strategies with different fine-tune epochs, one is to fix the parameters of the  backbone and fine-tune the FC layers, and the other one is to fine-tune the parameters of the whole network.  We can observe that the first fine-tuning strategy can achieve better performance in most settings, and the accuracy starts to converge at around 60 epochs, and increasing the number of fine-tuning epochs obtain slight improvement. Note that fine-tuning has a very high latency at inference time because the fine-tuning stage needs to calculate the gradient of the parameters in each epoch. Using a smaller backbone or classifier can increase inference speed, but the accuracy will drop accordingly. In contrast, our method obtains higher classification accuracy and faster inference speed.

    %================ table: fine-grained ================%
    \begin{table*}[t] \centering
    \renewcommand{\arraystretch}{1.2}
    \caption{ \label{table:fine_grained} Comparisons of the 5way-1shot-15query classification results (accuracy \%) under fine-grained setting on ShapeNet70-FS with DGCNN~\cite{wang2019dynamic} as the backbone.}
    \setlength{\tabcolsep}{1mm}{
    \begin{footnotesize}
    \scalebox{0.9}{
    \begin{tabular}{ccccccccc}
    \hline
    \makecell[c]{Methods}
    & Airline   & Jet    & Fighter 
    & \makecell[c]{Swept \\ Wing }   
    & \makecell[c]{Propeller\\Plane}
    & Bomber    & \makecell[c]{Delta\\Wing }
    & Mean 
    \\\hline 
    Prototypical Net~\cite{snell2017prototypical}
    & 49.59  & 25.21  & 27.64       & \tb{41.30}  
    & {35.76}  & 20.73  & 46.24       & {35.22} \\
    Relation Net~\cite{sung2018learning} 
    & 49.28       & 24.55  & 30.36       & \underline{41.03} 
    & 28.26       & 24.05  & 44.79       & 34.63   \\
    FSLGNN~\cite{garcia2017few}  
    & 48.53       & 23.56  & 21.01       & 31.75 
    & 24.29       & 23.58  & \tb{58.49}  & 33.08  \\
    Meta-learner~\cite{SachinRavi2017OptimizationAA}  
    & 29.76       & 21.92  & 30.94       & 27.83
    & 27.02       & 24.96  & 30.22       & 27.53 \\
    MAML~\cite{finn2017model}
    & \tb{56.03}  & \tb{26.65}  & \underline{34.37}  & 21.90  
    & 16.43       & {25.30}  & 15.71       & 28.12 \\
    MetaOptNet~\cite{lee2019meta} 
    & 46.81      & 24.72  & 29.05      & 39.74 
    & 24.42      & 25.17  & 38.23      & 32.60  \\\hline
    Sharma \etal~\cite{sharma2020self} 
    & 46.86      &23.75  & 30.21      & 35.98 
    & 35.98      & \underline{27.62}  & 40.96      & 34.37  \\
    Point-BERT~\cite{yu2021point} 
    & 50.05      & \tb{26.65}  & 32.82      & 39.36 
    & \tb{44.47}      & 22.71  & 44.44      & \underline{37.23}  \\\hline
    
    \multicolumn{1}{c}{\tb{Ours}} 
    & \underline{50.32 }     & \underline{{25.61}}   & \tb{{39.52}} & {40.13} 
    & \underline{{43.25}} & \tb{{30.92}}   & \underline{{47.06}} & \tb{{39.54}}  \\
    \hline
    \end{tabular}} 
    \end{footnotesize}}
    \end{table*}
    %================ table: fine-grained ================%

    %================ table: vanilla classification ================% 
    \begin{table*}[thb] \centering
    \renewcommand{\arraystretch}{1.3}
    \caption{\label{table:vanilla_classfication} { {The classification results of the SOTA methods, including PointNet++~\cite{qi2017pointnet++}, RSCNN~\cite{liu2019relation}, PointMLP~\cite{pointmlp}, equipped with the SPF and SCI+ module on ModelNet40 and ScanObjectNN benchmarks.}}}
    \setlength{\tabcolsep}{2mm}{
    \begin{footnotesize}
    \scalebox{0.85}{
    \begin{tabular}{c ccc ccc}
    \hline
    \multicolumn{1}{c }{\multirow{2}{*}
    {\makecell[c]{ {Method}}}} & 
    \multicolumn{3}{c }{ {ModelNet40}} & 
    \multicolumn{3}{c }{ {ScanObjectNN}} \\ \cline{2-7} 
    &   {Vanilla} &  {W/ SPF} &  {W/ SCI+} 
    &  {Vanilla} &  {W/ SPF} &  {W/ SCI+}   \\ \hline
    \multirow{1}{*}{ {PointNet2~\cite{qi2017pointnet++}}}
    & {92.66}	& {92.84}  {\scriptsize{(+0.18)}}	& {92.86}  {\scriptsize{(+0.20)}}	 & {78.24}	 & {79.64}  {\scriptsize{(+1.40)}}	& {79.25}  {\scriptsize{(+1.01)}} \\
   \multirow{1}{*}{ {RSCNN~\cite{liu2019relation}}}
    & {92.78}	& {93.01}  {\scriptsize{(+0.23)}}	& {92.95}  {\scriptsize{(+0.13)}}	 & {76.52}	 & {77.65}  {\scriptsize{(+1.13)}}	& {77.23}  {\scriptsize{(+0.69)}} \\
     \multirow{1}{*}{ {PointMLP~\cite{pointmlp}}}
   & {93.47}	& {93.63}   {\scriptsize{(+0.16)}}	& {93.59 }  {\scriptsize{(+0.12)}}	 & {85.39}	 & {86.00}   {\scriptsize{(+0.63)}}	& {85.87}  {\scriptsize{(+0.48)}} \\
    \hline
    \end{tabular}}
    \end{footnotesize}}  
    \end{table*}
    %================ table: vanilla classification ================%
    
    \subsection{ The SPF and SCI+ Module in Regular Classification}
     {We adopt the proposed SPF and SCI+ modules on three SOTA  methods~\cite{qi2017pointnet++,liu2019relation,pointmlp}, respectively, to study the improvement carried by these designs on regular point cloud classification task. Specifically, we add the SPF module before the FC classifier layer to refine the global feature. For the SCI+ module, we extended each channel of the global feature using 1*1 Conv directly (because there is no task-aware embedding in the general point cloud classification), and took the attention mechanism introduced in Sec 4.3 to adjust the output global feature. }
    
     {The classification results reported in Table R11 indicate that both the SPF and SCI+ module can improve the SOTA methods’ general point cloud classification performance, especially for the real-world dataset—ScanobjectNN (by about 0.5$\%$-1.5$\%$). We also can observe that the SPF module provides more performance gains than the SCI+ modules. According to this experiment result, we could conclude that the proposed SPF and SCI+ can help not only the few-shot point cloud task but also the general point cloud task to learn more representative features. }
    
     {It is also worth noting that the proposed SPF and SCI+ bring more significant performance gains under few-shot settings, i.e. nearly 1-2$\%$ as compared with baseline methods (refer to Table 7 in the manuscript), which is obviously higher than the gain (i.e. about 0.1-1\%) achieved on general classification settings. This is due to that the proposed SPF and SCI+ modules can improve the poor representative capability  of the network when samples are few. On the other hand, when samples are sufficient, the learned features will become more representative for each category, so the proposed modules have limited effect to improve the originally representative features. Therefore, based on the above analysis, we conclude that our proposed approach is more suitable for the few-shot point cloud classification task.}
    
    \vspace{-10pt}
    \subsection{Fine-Grained Few-Shot Classification}
    \vspace{-5pt}
    We also study the proposed network under the fine-grained classification setting to evaluate its ability to distinguish similar categories. We first train the baselines and our method on the training set of ShapeNet70-FS, and then test them on seven subcategories of “Airplane” in the testing set with a 5way-1shot-15query setting. The results are listed in Table~\ref{table:fine_grained}. One can see that our proposed network obtains better performance than other baselines for most subcategories and improves the mean accuracy by more than 2\%.
    
    \vspace{-5pt}
    \subsection{More t-SNE Visualisation}
\begin{figure}[thb]
    	\begin{center}
            \includegraphics[width=1\linewidth]{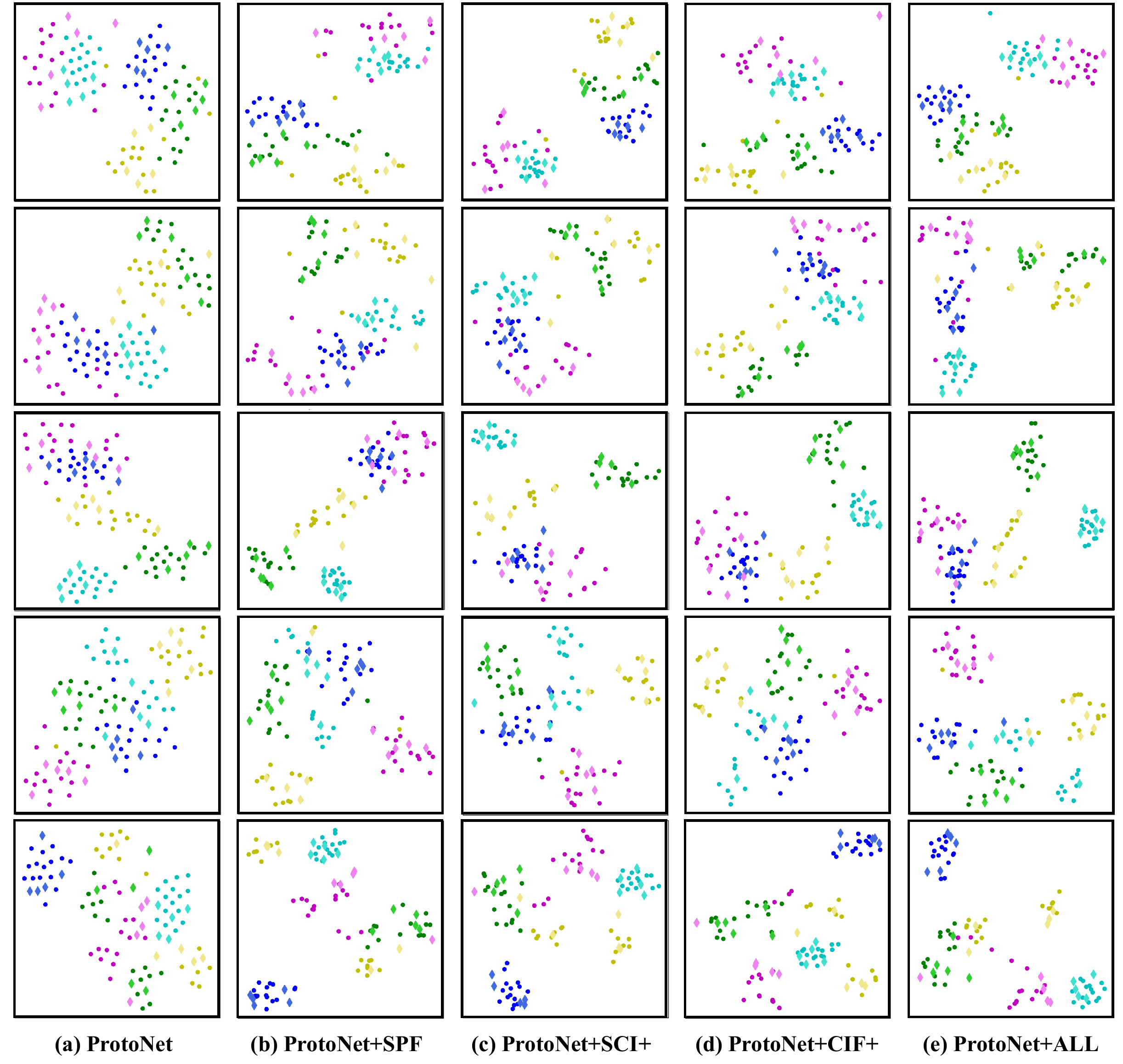} \vspace{-10pt}
    	\end{center} 
    	\caption{The t-SNE visualization results. $\Diamond$ stands for the support features,  $\bullet$ represents the query features.  We use the same color to represent the same class.  }
    	\label{fig:tsne2} 
    	\vspace{-10pt}
    \end{figure}

    \vspace{-5pt}
    Similar to Section 5.6.1, we provide more t-SNE visualization results in Figure~\ref{fig:tsne2}. We sample five episodes in the test split of Shapenet70-FS  under the 5-way 5-shot 15-query setting with DGCNN as the embedding backbone. Note that the learning support and query features of ProtoNet are dispersed with larger distribution shifts. After using the proposed modules, the features tend to move to the center to better differentiate from other classes, and the distribution shift between support and query sets is mitigated.
     
    \subsection{More Visualisation on  Real-World 3D Scan}
\begin{figure}[thb]
    	\begin{center}
            \includegraphics[width=1\linewidth]{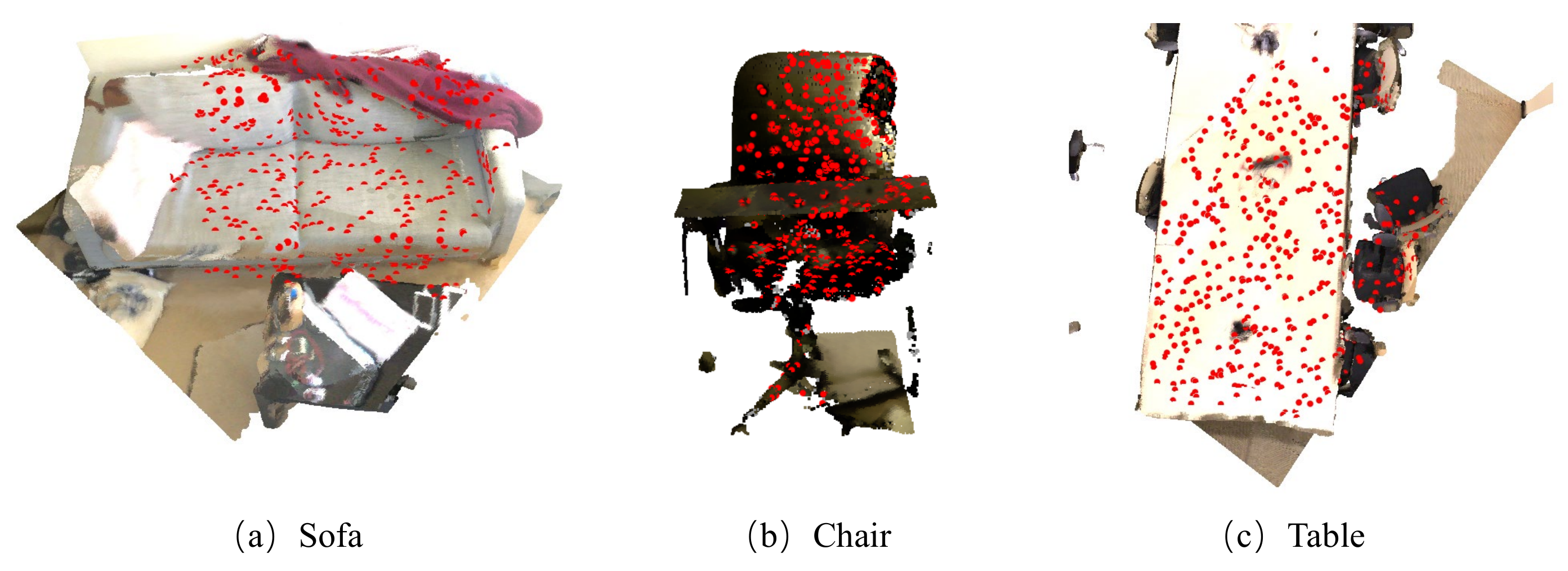} \vspace{-10pt}
    	\end{center} 
    	\caption{Visualization of selected salient points by the SPF in real-world 3D scan data }
    	\label{fig:real-world} 
    	\vspace{-10pt}
    \end{figure}

    This section provides more visualisation of elected salient points by the SPF in real-world 3D scan data, as shown in Fig.~\ref{fig:real-world}. We can see that the most selected salient points locate in the foreground object, such the Sofa, Chair and Table.
     
     %================ fig: real-world  ================%
    %================ fig: real-world  ================%
     
    %================ fig: tsne2  ================%
    %================ fig: tsne2  ================%

\section{Dataset Split}
    \label{dataset_split}
    In this section, we offer the splitting details (class name and number) and statistics of each benchmark dataset, which are listed in Table~\ref{table:statict_Strain}, ~\ref{table:statict_Stest},~\ref{table:statict_Mtrain}, ~\ref{table:statict_ScanObject}, respectively. 

 \vspace{-5pt}
 
   %================ table: ScanObjectNN-FS  ================%
    \begin{table}[htp]
        \centering
        \renewcommand{\arraystretch}{1.1}
         \caption{\label{table:statict_ScanObject} Class names and numbers of each split of ScanObjectNN-FS.}  
    \setlength{\tabcolsep}{3mm}{
        \begin{footnotesize}
         \scalebox{0.9}{
    \begin{tabular}{cc|cc|cc}
    \hline
    \multicolumn{2}{c|}{Split=0}& \multicolumn{2}{c|}{Split=1} &\multicolumn{2}{c}{Split=2}  
    \\ \hline
    Class     & Num & Class     & Num & Class   & Num   \\ \hline
    Shelf	&1325 	&Chair	&1975  &Cabinet	&1716	\\
    Door	&1102   &Sofa	&1268  &Table	&1192	 	\\
    Bin	&993  	&Desk	&742   &Display	&882	\\
    Box	&539    &Bed	&674   &Sink	&589	 	\\
    Bag	&381  	&Pillow	&510   &Toilet	&410	\\
   \hline
    \end{tabular}}
    \end{footnotesize}} 
    \vspace{-10pt}
    \end{table}
   %================ table: ScanObjectNN-FS  ================%

   %================ table: ModelNet-FS  ================%
    \begin{table}[htp]
    \renewcommand{\arraystretch}{1.1}
    \centering
    \caption{\label{table:statict_Mtrain} The training and testing classes of ModelNet40-FS dataset.} 
    \setlength{\tabcolsep}{1mm}{
    \begin{footnotesize}
    \scalebox{0.9}{
    \begin{tabular}{cc|cc|cc|cc}
    \hline
    \multicolumn{6}{c|}{Training Set}  &\multicolumn{2}{c}{Testing Set}    \\ \hline
    Class     & Num & Class     & Num & Class   & Num & Class & Num \\ \hline
    chair     & 989   & car       & 297        & radio  & 124  &bookshelf & 672  \\  
    sofa      & 780   & desk      & 286        & xbox   & 123   &vase      & 575    \\
    airplane  & 725   & dresser   & 286     & bathtub   & 156  & bottle    & 435   \\
    bed       & 615   & glass\_box & 271     & lamp    & 144    & piano     & 331   \\
    monitor   & 565   & guitar     & 255     & stairs  & 144    & night\_stand & 286\\
    table     & 492   & bench      & 193    & door     & 129   & range\_hood  & 215\\ 
    toilet    & 444   & cone       & 187    & stool  & 110     & flower\_pot & 169\\
    mantel    & 384   & tent       & 183   & wardrobe& 107    & keyboard    & 165\\
    tv\_stand & 367   & laptop     & 169   & cup    & 99       & sink   & 148\\
    plant     & 339   & curtain    & 157   & bowl   & 84       & person & 108\\
   \hline
    \end{tabular}}
    \end{footnotesize}}   
    \end{table}
   %================ table: ModelNet-FS  ================%

   %================ table: ShapeNet-FS  ================%
    \begin{table}[htp]
    \renewcommand{\arraystretch}{1.1}
        \centering
        \caption{\label{table:statict_Strain} The training classes of ShapeNet70-FS dataset. "ID" corresponds to WordNet synset offset. }  
    \setlength{\tabcolsep}{1mm}{
        \begin{footnotesize}
         \scalebox{0.9}{
    \begin{tabular}{ccc|ccc}
    \hline
    \multicolumn{6}{c}{Training Set}                                                                           \\ \hline
    ID        & Class          & Num  & ID       & Class             & Num   \\ \hline
    04256520  & sofa           & 1520 & 04037443 & race car          & 323    \\
    03179701  & desk           & 1226 & 20000011 & garage cabinet    & 307    \\
    04401088  & phone          & 1089 & 03948459 & handgun           & 307    \\
    02738535  & armchair       & 1051 & 04285965 & sport utility     & 300    \\
    02924116  & bus            & 939  & 03928116 & piano             & 239    \\
    02808440  & bathtub        & 856  & 02818832 & bed               & 233    \\
    02992529  & radiotelephone & 831  & 04330267 & stove             & 218    \\
    03891251  & park bench     & 823  & 03100240 & convertible       & 208    \\
    03063968  & coffee table   & 763  & 04285008 & sports car        & 197   \\
    20000027  & club chair     & 748  & 02880940 & bowl             & 186      \\
    02858304  & boat           & 741  & 03141065 & cruiser          & 181      \\
    04250224  & sniper rifle   & 717  & 02961451 & carbine          & 172      \\
    03046257  & clock          & 651  & 04004475 & printer          & 166      \\
    03991062  & pot            & 602  & 03761084 & microwave        & 152      \\
    03593526  & jar            & 596  & 04225987 & skateboard       & 152      \\
    03237340  & dresser        & 482  & 04460130 & tower            & 133      \\ 
    04380533  & table lamp     & 464  & 20000020 & cantilever chair & 125      \\
    03642806 & laptop            & 460 & 02801938 & basket           & 113 \\
    04166281 & sedan             & 429 & 02814533 & beach wagon      & 108 \\
    03624134 & knife             & 424 & 02946921 & can              & 108 \\
    20000037 & rectangular table & 421 & 03938244 & pillow           & 96  \\
    03119396 & coupe             & 418 & 03594945 & jeep             & 95  \\
    04373704 & swivel chair      & 398 & 03207941 & dishwasher       & 93  \\
    20000010 & desk cabinet      & 356 & 04099429 & rocket           & 85  \\
    03790512 & motorcycle        & 337 & 02773838 & bag              & 83  \\
    \hline
    \end{tabular}}
    \end{footnotesize}} 
    \end{table}
    %================ table: ShapeNet-FS  ================%
    
    %================ table: ShapeNet-FS  ================%
    \begin{table}[htp]
    \renewcommand{\arraystretch}{1.1}
        \centering
        \caption{\label{table:statict_Stest} The testing classes of ShapeNet70-FS dataset. "ID" corresponds to WordNet synset offset.}  
    \setlength{\tabcolsep}{2mm}{
        \begin{footnotesize}
         \scalebox{0.9}{
    \begin{tabular}{ccc|ccc}
    \hline
    \multicolumn{6}{c}{Testing Set}                                                                            \\ \hline
    ID        & Class          & Num  & ID       & Class             & Num  \\ \hline
    03211117  & display        & 1093  & 03337140 & file cabinet      & 298         \\
    02690373  & airline        & 1054  & 20000001 & swept wing        & 271         \\
    03467517  & guitar         & 797   & 03797390 & mug               & 214         \\
    03325088  & faucet         & 744   & 04554684 & washer           & 169      \\
    03595860  & jet            & 675   & 03513137 & helmet           & 162          \\
    03335030  & fighter        & 597   & 04012084 & propeller plane  & 137          \\
    02876657 & bottle          & 498   & 02867715 & bomber           & 130       \\
    02871439 & bookshelf         & 452 & 03174079 & delta wing       & 121  \\
    04468005 & train             & 389 & 02942699 & camera           & 113  \\
    02747177 & ashcan            & 343 & 03710193 & mailbox          & 94   \\
    \hline
    \end{tabular}}
    \end{footnotesize}}  \vspace{-10pt}
    \end{table}
   %================ table: ShapeNet-FS  ================%
   
% \clearpage
% % BibTeX users please use one of
% \bibliography{Mendeley}   % name your BibTeX data base

\end{document}